\documentclass[letterpaper]{article} 
\usepackage{aaai2027}  
\usepackage[hyphens]{url}  
\usepackage{graphicx} 
\usepackage{natbib}  
\usepackage{caption} 
\usepackage{algorithm}
\usepackage[noend]{algpseudocode}
\usepackage{amsmath}
\usepackage{amssymb}
\usepackage{amsthm}
\usepackage{tikz}
\usepackage{multirow}
\usepackage{cuted}
\usepackage{capt-of}
\usetikzlibrary{positioning,arrows.meta,fit,backgrounds,shapes.geometric,calc}

\usepackage{booktabs}
\usepackage{array}
\usepackage{makecell}
\usepackage{enumitem}
\usepackage[most]{tcolorbox}
\usepackage{pifont}
\newcommand{\cmark}{\ding{51}}
\newcommand{\xmark}{\ding{55}}
\title{When Absence Is Evidence: Evaluating Completeness-Sensitive Negative Reasoning in Large Language Models}
\author{
    Byoungjae Min\textsuperscript{\rm 1},
    Kennedy Edemacu\textsuperscript{\rm 2},
    Sae-Hong Cho\textsuperscript{\rm 3},\\
    Yoonhyuk Choi\textsuperscript{\rm 4},
    Beakcheol Jang\textsuperscript{\rm 5},
    Jong Wook Kim\textsuperscript{\rm 1}
}
\affiliations{
    \textsuperscript{\rm 1}Department of Computer Science, Sangmyung University, Seoul, Republic of Korea\\
    \textsuperscript{\rm 2}College of Staten Island, The City University of New York, New York, NY, USA\\
    \textsuperscript{\rm 3}School of Computer Engineering, Hansung University, Seoul, Republic of Korea\\
    \textsuperscript{\rm 4}Sookmyung Women's University, Seoul, Republic of Korea\\
   \textsuperscript{\rm 5}Graduate School of Information, Yonsei University, Seoul, Republic of Korea\\
    jkim@smu.ac.kr
}

\begin{document}
\nocopyright
\maketitle

\begin{abstract}
Large language models (LLMs) are often asked whether something is absent
from a record, list, or retrieved context. Yet non-observation licenses a
negative answer only when evidence completely covers the query scope;
otherwise, the answer should remain unknown. We call this
completeness-sensitive negative reasoning. We introduce CROWN-QA,
comprising CROWN-Synth, a controlled paired core that fixes the question
and observed facts while varying only query-relative coverage, and
CROWN-Real, a real-document contrast-set evaluation with controlled
coverage variants. Across three LLM families, models show unstable
closure judgments and substantial over-closure, failing to reliably
distinguish a justified negative answer
(\textsc{Certified-Negative}) from insufficient evidence
(\textsc{Unknown}). The dominant CROWN-Synth failure is asymmetric:
models often recognize implicitly complete evidence yet treat implicitly
partial evidence as query-covering. Prompting redistributes errors between
over- and under-closure rather than consistently resolving them.
Structured certificate elicitation traces many errors to
evidence-coverage mischaracterization. CROWN-Real shows that the core
partial-coverage asymmetry persists on real-document content, while its
strength and the balance between over- and under-closure vary by model,
prompt, and source.
\end{abstract}


\section{Introduction}

Large language models (LLMs) are commonly evaluated by their ability to
find and use positive evidence. In retrieval-augmented generation (RAG),
for example, evaluation often asks whether retrieved context is relevant,
sufficient, and faithfully used to answer the user query~\cite{lewis2020rag,es2024ragas,saadfalcon2024ares,niu2024ragtruth,chen2024rgb,yang2024crag,joren2025sufficient}.
In real settings, however, many questions---including high-stakes
ones---require reasoning about absence: whether a condition is absent from
a contraindication list, an applicant is absent from an exclusion list, or
a proceedings index contains no matching title.

The key challenge is that absence is evidential only when coverage is both
complete and query-covering. Figure~\ref{fig:intro-example} holds the query
and listed titles fixed while varying coverage status and scope. The complete conference-wide index (center) licenses
\textsc{Certified-Negative}, whereas both the partial search results
(left) and the complete main-track index (right) remain
\textsc{Unknown}. Thus, the core problem is not detecting absence, but
deciding when absence licenses negation.

\begin{figure}[t]
\centering
\small
\setlength{\tabcolsep}{6pt}
\renewcommand{\arraystretch}{0.75}

\begin{minipage}{0.95\columnwidth}
\textbf{Query:} Did any MLConf 2024 paper, across all tracks, have a
title containing ``calibrated retrieval''?\\[-1pt]
\textbf{Observed titles:} The listed titles are identical across all three
variants; none contains the queried phrase.
\end{minipage}

\vspace{2pt}
\begin{tabular}{p{0.27\columnwidth} p{0.32\columnwidth} p{0.27\columnwidth}}
\toprule
\multicolumn{1}{c}{\shortstack{\textbf{Partial}\\\textit{all tracks}}} &
\multicolumn{1}{c}{\shortstack{\textbf{Complete}\\\textit{all tracks}}} &
\multicolumn{1}{c}{\shortstack{\textbf{Complete}\\\textit{main track only}}} \\
\midrule
Top-20 keyword-search results. &
Official conference-wide title index. &
Official main-track title index. \\
\midrule
\multicolumn{1}{c}{\textsc{Unknown}} &
\multicolumn{1}{c}{\textsc{Certified-Negative}} &
\multicolumn{1}{c}{\textsc{Unknown}} \\
\bottomrule
\end{tabular}

\caption{Matched variants with identical titles. Only the complete
conference-wide index covers the query; the partial results and the complete
main-track index remain \textsc{Unknown}.}
\label{fig:intro-example}
\end{figure}

\begin{figure}[t]
\centering
\includegraphics[width=0.99\columnwidth]{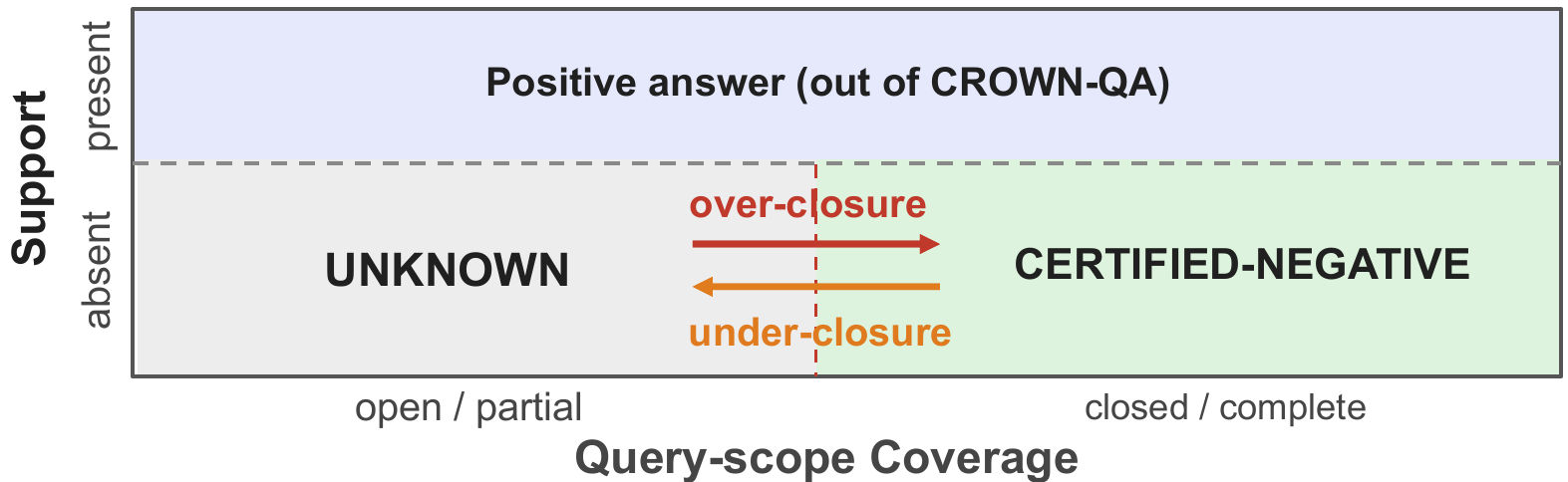}
\caption{Conceptual outcome space. Positive-support cases are ordinary QA and are outside CROWN-QA; CROWN-QA evaluates the no-support branch, where query-scope coverage determines \textsc{Certified-Negative} versus \textsc{Unknown}.}
\label{fig:axes}
\end{figure}

Completeness is query-relative: evidence may be complete for its own
scope yet fail to cover the query scope. As Figure~\ref{fig:axes} shows, support and query-scope coverage are
separate axes: a justified ``no'' requires absent support and complete
query-covering evidence. CROWN-QA fixes the former by
construction and evaluates only the latter. This scope-containment view
is related to query--knowledge relevance in
RAG~\cite{li2024queryknowledge} and gives a query-relative form of the
open- versus closed-world distinction: non-closing evidence leaves an
unobserved fact unknown, whereas complete query-covering evidence
licenses a negative
answer~\cite{razniewski2024completeness}. This
distinction is rarely isolated in natural-language LLM evaluation.

Prior work studies abstention, unanswerability, ambiguous information
needs, and knowledge boundaries~\cite{peng2025unanswerability,
kirichenko2025abstention,madhusudhan2025notanswer,zhang2024clamber,
chen2025knowledgeboundary}, retrieval robustness, context sufficiency,
evidence-based QA, and factuality~\cite{yu2024chainofnote,
joren2025sufficient,glockner2025neoqa,bayat2025factbench}, as well as
omitted information~\cite{fu2025absencebench}, false
premises~\cite{shafiei2025multihoax}, and
negation~\cite{garcia2023not}. These settings ask whether to answer,
refuse, or identify missing or misleading information; they do not hold
the query and observed facts fixed while varying whether coverage
contains the query scope. CROWN-Synth isolates this factor after fixing
support as absent: the same unsupported fact remains
\textsc{Unknown} under non-closing evidence but becomes
\textsc{Certified-Negative} under complete query-covering evidence.
Thus, always abstaining fails on the latter, whereas always answering
``no'' fails on the former.

We introduce CROWN-QA (Completeness Reasoning Over What is Not in
Question Answering), combining a controlled paired core, real-document
contrast sets, and diagnostic certificates for query scope, evidence
scope, and closure judgment. The contributions of this paper are as
follows.

\begin{itemize}[leftmargin=9pt]

\item \textbf{Problem and formalization.}
We define completeness-sensitive negative reasoning as an
absence-conditioned QA task: \textsc{Certified-Negative} is justified
only when evidence completely covers the query scope; otherwise, the
answer is \textsc{Unknown}.

\item \textbf{Benchmark and diagnostics.}
We introduce CROWN-QA: CROWN-Synth, a controlled paired core with
L1--L4 regimes, Scope-Mismatch cases, and paired completeness
sensitivity; and CROWN-Real, a real-document contrast-set evaluation.
Directional metrics and structured certificates support diagnosis. To
our knowledge, CROWN-Synth is the first controlled natural-language QA
benchmark to isolate query-relative coverage as the label-changing
intervention.

\item \textbf{Empirical findings and diagnostic analysis.}
Across three LLM families, CROWN-Synth shows strong over-closure,
especially on implicitly partial evidence, while prompting shifts errors
between over- and under-closure. CROWN-Real reproduces the
partial-coverage gap on real-document content, but its size varies by
model, prompt, and source. Certificate errors most often first appear in
the reported evidence-coverage field.

\end{itemize}

\section{Related Work}

\begin{table}[t]
\centering
\scriptsize
\setlength{\tabcolsep}{4pt}
\renewcommand{\arraystretch}{0.95}
\resizebox{\columnwidth}{!}{%
\begin{tabular}{l ccccc}
\hline
Work & (a) & (b) & (c) & (d) & (e) \\
\hline
Sufficient Ctx.~\cite{joren2025sufficient}        & \xmark & \xmark & \xmark & \xmark & \xmark \\
UAEval4RAG~\cite{peng2025unanswerability}          & \xmark & \xmark & \xmark & \xmark & \xmark \\
AbstentionBench~\cite{kirichenko2025abstention}    & \xmark & \xmark & \xmark & \xmark & \xmark \\
AbsenceBench~\cite{fu2025absencebench}             & \cmark$^\ast$ & \xmark & \xmark & \xmark & \xmark \\
Negation~\cite{garcia2023not}                      & \cmark$^\ast$ & \xmark & \xmark & \xmark & \xmark \\
KB completeness/negation~\cite{razniewski2024completeness}   & \cmark & \cmark & \xmark & \cmark$^\dagger$ & --- \\
\textbf{CROWN-QA}                                  & \cmark & \cmark & \cmark & \cmark & \cmark \\
\hline
\end{tabular}}
\caption{Positioning of CROWN-QA. Dimensions: (a)~absence or explicit negation as the primary semantic target; (b)~query-relative set coverage or completeness explicitly modeled;
(c)~same-question, same-fact paired coverage interventions;
(d)~query--coverage Scope-Mismatch; and (e)~negative licensing
distinguished from uncertainty in NL QA. $^\ast$ These works evaluate
omission or explicit negation, but not whether missing support licenses
a negative answer. $^\dagger$ KB completeness represents scope
symbolically rather than inferring it from language; \emph{---} indicates
that (e) does not apply to symbolic KB semantics.}
\label{tab:delta}
\end{table}

\noindent\textbf{RAG sufficiency and unanswerability. }
RAG grounds generation in retrieved
context~\cite{lewis2020rag}, and evaluations measure relevance,
faithfulness, hallucination, sufficiency, and
unanswerability~\cite{saadfalcon2024ares,niu2024ragtruth,
joren2025sufficient,peng2025unanswerability}. Related benchmarks examine
missing or misleading evidence~\cite{glockner2025neoqa},
query--knowledge relevance~\cite{li2024queryknowledge}, and broader
abstention cases~\cite{kirichenko2025abstention}.  These works primarily
evaluate answer support, context sufficiency, or abstention.

\noindent\textbf{Absence and negation in LLMs.}
Recent work shows that LLMs struggle with omitted information and
explicit negation~\cite{fu2025absencebench,garcia2023not}.
Omitted-content benchmarks ask what is missing, while negation benchmarks
test linguistic negation. 

\noindent\textbf{Completeness in knowledge bases.}
Database and knowledge-base research distinguishes closed-world
assumptions, where missing facts are treated as false, from open-world
assumptions, where they remain
unknown~\cite{razniewski2024completeness}. Partial
completeness and completeness assertions encode coverage through formal
semantics or metadata~\cite{razniewski2015extent,
darari2013completeness}. 

\noindent\textbf{Selective prediction and knowledge boundaries.}
Selective prediction trades coverage for risk, while conformal methods
provide uncertainty sets or risk-control
guarantees~\cite{elyaniv2010foundations,geifman2017selective,
angelopoulos2021gentle}; related LLM work studies model self-knowledge,
unanswerability, and clarification under
ambiguity~\cite{kadavath2022know,yin2023know,rajpurkar2018know,
cole2023selectively,madhusudhan2025notanswer,zhang2024clamber}.
These lines manage unreliable answers through abstention, uncertainty
sets, or clarification.

\noindent\textbf{Positioning.}
Across these lines, the absence of support is handled as a reason to
abstain, hedge, or clarify; none of them tests when absence itself
licenses a negative answer. CROWN-QA targets exactly this distinction
(Table~\ref{tab:delta}): CROWN-Synth isolates it by fixing the question
and observed facts and varying query-relative coverage alone, and
CROWN-Real tests whether the same distinction transfers to real
documents.

\section{Task Definition}

\subsection{Absence-Conditioned Completeness Judgment}
Let \(q\) be a natural-language question and let
\(E=\{e_1,\ldots,e_n\}\) denote the evidence context. The context may contain factual statements and coverage information,
expressed explicitly or through source semantics, that specifies the
scope described by \(E\) and whether its coverage is complete, partial,
or unspecified. Such a scope
may involve a time period, entity, attribute, list, database field, or
document collection. We denote the scope requested by the question as
\(S(q)\) and the coverage scope asserted by the evidence as \(S(E)\).

CROWN-QA is conditioned on the absence of positive support: the queried
fact is not supported by the observed factual content of \(E\). The task
is therefore not to find a positive answer, but to decide whether this
observed absence is licensed as a negative answer by the coverage
information. The model must output one of two labels: 
\[ \mathcal{Y}=\{\textsc{Certified-Negative},\textsc{Unknown}\}. \]
\textsc{Certified-Negative} is correct when the observed absence is
licensed by complete coverage of the query scope; ``certified'' is
limited to the stated evidence coverage. \textsc{Unknown} is correct when coverage is partial, sampled,
unspecified, ambiguous, or complete only for a scope that does not cover
\(S(q)\).

Formally, let \(c=\mathrm{Comp}(E,S(q))\in\{0,1\}\) indicate whether
\(E\) closes \(S(q)\). Here \(c=1\) only when \(E\) establishes complete coverage of \(S(E)\)
and \(S(q)\subseteq S(E)\). Since positive support is absent by construction, the
target label is
\[
y^*(q,E)=
\begin{cases}
\textsc{Certified-Negative}, & c=1,\\
\textsc{Unknown}, & c=0.
\end{cases}
\]
Here \(c=1\) only when \(E\) establishes complete coverage of its asserted
scope \(S(E)\) and \(S(q)\subseteq S(E)\). This containment relation is
semantic rather than lexical; a complete source for a different time,
entity, attribute, population, or collection has \(c=0\).

CROWN-QA excludes positive-evidence retrieval, already studied in QA and
RAG, and isolates the downstream licensing step: once support is absent,
does coverage justify a negative answer?

\subsection{Failure Modes}

Although the gold label is deterministic, the model must infer
query-relative coverage from natural language. Let
\(\hat y\in\mathcal{Y}\) denote the model prediction and let \(y^*\)
denote the gold label. We focus on two directional errors that aggregate
accuracy can obscure.

\textbf{Over-Closure.}
Over-closure occurs when a model outputs
\textsc{Certified-Negative} even though the evidence is non-closing for
the query scope:
\[
\hat{y}=\textsc{Certified-Negative}, \qquad
y^*=\textsc{Unknown}.
\]
This corresponds to treating observed absence as evidence of absence
without licensed coverage.

\textbf{Under-Closure.}
Under-closure occurs when a model outputs \textsc{Unknown} even though
complete query-covering evidence licenses a certified negative answer:
\[
\hat{y}=\textsc{Unknown}, \qquad
y^*=\textsc{Certified-Negative}.
\]
This corresponds to unnecessary abstention despite licensed negative
evidence.

\subsection{Metrics}

We first report the over-closure rate (OCR) and under-closure rate (UCR):
{\small
\[
\begin{aligned}
\mathrm{OCR}
&=\Pr[\hat y=\textsc{Certified-Negative}\mid
       y^*=\textsc{Unknown}],\\
\mathrm{UCR}
&=\Pr[\hat y=\textsc{Unknown}\mid
       y^*=\textsc{Certified-Negative}].
\end{aligned}
\]
}

\noindent OCR measures false certification under non-closing evidence, whereas UCR
measures false abstention under complete query-covering evidence.

We report class-balanced accuracy, denoted \(\mathrm{Acc}\), which
averages the correct rates for the two gold labels:
{\small
\[
\begin{aligned}
\mathrm{Acc}=\tfrac{1}{2}\big(&
\Pr[\hat y=\textsc{Certified-Negative}\mid
    y^*=\textsc{Certified-Negative}]\\[-1pt]
&+\Pr[\hat y=\textsc{Unknown}\mid
    y^*=\textsc{Unknown}]
\big).
\end{aligned}
\]
}

\noindent We use this definition for both CROWN-Synth and CROWN-Real, so the two
gold labels receive equal weight despite their different label ratios.

For CROWN-Synth, we additionally report paired completeness sensitivity
(CS) on same-question, same-fact pairs. Let \(x_c=(q,E_c)\) denote the
complete query-covering member with gold label
\textsc{Certified-Negative}, and let \(x_o=(q,E_o)\) denote its matched
non-closing member with gold label \textsc{Unknown}. CS is the fraction
of pairs answered correctly in both directions:
{\small
\[
\mathrm{CS}=\Pr[\hat{y}(x_c)=\textsc{Certified-Negative}\ \wedge\
\hat{y}(x_o)=\textsc{Unknown}].
\]
}

\noindent High CS indicates that the model changes its answer in the intended
direction when only query-relative coverage changes.

\section{CROWN-QA Benchmark}

CROWN-QA combines two forms of coverage control. CROWN-Synth uses exact
same-question, same-fact pairs, whereas CROWN-Real uses A/B/C contrast
sets grounded in real documents. All examples are absence-conditioned;
Table~\ref{tab:dataset-stats} summarizes both components.

\begin{table}[t]
\centering
\scriptsize
\setlength{\tabcolsep}{2.8pt}
\renewcommand{\arraystretch}{0.92}
\begin{tabular}{@{}p{1.1cm}p{3.4cm}p{3.4cm}@{}}
\toprule
\textbf{Property} & \textbf{CROWN-Synth} & \textbf{CROWN-Real} \\
\midrule

Source
& Synthetic worlds
& ACL Anthology proceedings;
  DailyMed drug labels \\

Domains
& 5
& 2 \\

Unit
& Matched pair (2 variants)
& Contrast set (A/B/C) \\

Scale
& 5{,}000 examples (2{,}500 pairs)
& 1{,}599 examples (533 sets) \\

CN:UNK
& 1:1
& 1:2 \\

Non-closing
& Partial / Scope-Mismatch (1:1)
& B: narrower complete;
  C: non-exhaustive (1:1) \\

Control
& Same question and observed facts
& Same question and target phrase;
  source content varies \\

Coverage
& \makecell[l]{L1--L4 coverage-expression\\regimes}
& \makecell[l]{A: query-covering complete;\\
               B: narrower complete;\\
               C: non-exhaustive} \\
\bottomrule
\end{tabular}

\caption{CROWN-QA design summary. All examples are
absence-conditioned. CROWN-Synth provides exact paired control,
whereas CROWN-Real tests transfer on real-document content with
controlled coverage variants.}
\label{tab:dataset-stats}
\end{table}

\subsection{CROWN-Synth: Controlled Paired Core}
CROWN-Synth is the controlled core of CROWN-QA. It is generated from
synthetic worlds in which observed factual content, query scope, evidence
coverage scope, and coverage status can be independently controlled.
This design isolates whether a model changes its answer because evidence
closes the query scope, rather than because the queried item or
surrounding factual content changes.

\textbf{Paired construction.}
For each base world, we generate a queried fact \(r\) and an observed
evidence set in which \(r\) is absent. We then create matched members
with the same question and identical observed factual content. The
query-covering member establishes complete coverage of \(S(q)\), yielding
\textsc{Certified-Negative}. Its matched non-closing member is partial,
sampled, unspecified, or complete only for a scope that does not contain
\(S(q)\), yielding \textsc{Unknown}. A worked pair is provided in
Appendix~A.

\textbf{Coverage regimes.}
A benchmark that always states ``this is the complete record'' can reduce
to keyword spotting. We therefore vary how coverage is expressed across
four regimes: (L1) explicit---``this is the complete list'';
(L2) paraphrased---``the official registry of all \(X\)'';
(L3) implicit---coverage is conveyed by a balanced inventory of four
complete and four partial source-type families, crossed with domain and
realized in a shared sentence frame, without explicit completeness
language; and (L4) adversarial---a complete source is
saliently mentioned but does not establish that the displayed evidence
closes the query scope. L1--L3 vary coverage explicitness, whereas L4
tests robustness to a non-licensing completeness statement. Coverage
regime and coverage relation are annotated independently, so
Scope-Mismatch cases occur at every L1--L4 regime. We report the two axes
separately to distinguish cue matching from scope reasoning.

\textbf{Scope-Mismatch cases.}
Complete--partial pairs, particularly under explicit coverage regimes,
may reward detecting whether a closure cue is present. We therefore add
cases in which the evidence is complete for a scope \(S(E)\) that does
not contain the query scope \(S(q)\). For example, a question may ask
whether any MLConf 2024 paper across all tracks has a matching title,
while the evidence provides a complete index of main-track papers only.
Although support is absent and the source is complete for its own scope,
\(S(q)\not\subseteq S(E)\), so the correct label is
\textsc{Unknown}. These cases expose policies that treat any completeness
cue as sufficient and test semantic scope containment rather than shallow
cue matching. Each item is annotated as a period-overlap mismatch,
hierarchy mismatch, population mismatch, or attribute mismatch.

\textbf{Domain templates.}
The benchmark covers domains in which negative answers are practically meaningful. Academic records test whether a requirement is absent from a complete student record. Grant eligibility tests whether an applicant is excluded under a complete list of exclusion criteria. Medical contraindications test whether a condition appears in a contraindication list declared complete for the relevant scope. Legal and policy rules test whether an exception or prohibition applies. Proceedings and catalogs test whether an item is absent from an official list declared complete for the relevant scope.

\subsection{CROWN-Real: Real-Document Transfer Set}

CROWN-Real tests whether the failure patterns isolated on CROWN-Synth
persist when coverage must be interpreted from real-document content
and source structure. It is grounded in two source families: NLP
proceedings from the ACL Anthology~\cite{aclanthology} and public drug labels from DailyMed~\cite{dailymed}.
Unlike the strict same-fact pairs in CROWN-Synth, CROWN-Real uses
three-member contrast sets that share a question and an exact target
phrase, while the displayed source content varies across members. The
target phrase is absent from every displayed context. Variant A provides
complete query-covering evidence, Variant B is complete for a narrower
scope, and Variant C is non-exhaustive.

For proceedings, A combines the full title indexes of two official
collections, B retains one complete collection, and C contains a
deterministic strict subset of titles from the same two collections.
For drug labels, A contains the full Warnings and Precautions section
with its source-native boundaries, B contains one complete numbered
subsection with its scope and closing boundary, and C uses the Highlights
material as an explicitly non-exhaustive control. Thus, titles, label
text, headings, and document boundaries come from real sources; coverage
is varied through controlled extraction and selection rather than
generated factual claims.

We verify by normalized exact matching that the queried phrase is absent
from every variant and exclude any item containing positive support.
Source identity and section or collection boundaries are retained so
that closure is licensed by the evidence shown to the model rather than
by hidden metadata. Medical C is analyzed only as an explicit-partial
control; implicit-partial transfer is evaluated on proceedings C.
Because CROWN-Real uses contrast sets rather than strict pairs, we report
class-balanced \(\mathrm{Acc}\) and variant-specific recalls instead of
paired completeness sensitivity.

\subsection{Quality Control}

Each CROWN-Synth pair must satisfy three constraints:
(i) the queried fact is absent from the observed factual content;
(ii) the question and observed factual content are identical across the
paired members; and (iii) only coverage status or scope changes the gold
label. Gold labels follow controlled metadata through
\(c=\mathrm{Comp}(E,S(q))\), rather than human annotation. We manually
inspect a stratified sample across domains, coverage regimes, and
coverage relations, using an LLM judge as a secondary consistency check.
We reject implicit-partial items that inadvertently establish exhaustive
coverage and Scope-Mismatch items solvable by an obvious non-overlapping
token difference, such as disjoint entities or years. 

For CROWN-Real, we verify by normalized exact matching that the target is
absent from every A/B/C member. We also check source identity, section or
collection boundaries, and whether the displayed evidence itself
establishes the intended A/B/C coverage relation. Items containing
positive support or requiring hidden metadata to justify the gold label
are excluded.

\section{Structured Scope-and-Coverage Elicitation}

To examine where closure errors arise, we evaluate a structured
elicitation condition using the same base LLM. Given a question \(q\)
and evidence context \(E\), the model returns a completeness certificate $C=(\hat S_q,\hat S_E,\hat c)$, where \(\hat S_q\) is the model-reported query scope, \(\hat S_E\)
describes the model-reported evidence coverage, including its scope and
completeness status, and \(\hat c\in\{0,1\}\) is the final coverage
judgment. All three fields are elicited jointly in one structured
response. Only \(\hat c\) is mapped to the scored label; the two scope
fields are retained for diagnostic analysis.

The model should set \(\hat c=1\) only when the coverage described by
\(\hat S_E\) is complete and contains the full scope described by
\(\hat S_q\). Accordingly, \(\hat c=0\) for partial, sampled,
incomplete, unspecified, ambiguous, or scope-mismatched coverage. This
is a semantic judgment rather than a literal comparison of
natural-language strings. Reporting the two scopes separately makes the query--coverage relation
explicit, particularly in Scope-Mismatch cases, rather than collapsing
the decision into a single sufficiency judgment.

For scoring, the Boolean field is mapped to the common two-label space:
\[
\hat{y}(C)=
\begin{cases}
\textsc{Certified-Negative}, & \hat{c}=1,\\
\textsc{Unknown}, & \hat{c}=0.
\end{cases}
\]
Because \(c\) is the gold coverage judgment, the directional error rates
for this condition are
\[
\mathrm{OCR}=\Pr[\hat{c}=1\mid c=0],
\qquad
\mathrm{UCR}=\Pr[\hat{c}=0\mid c=1].
\]
We compare \(\hat S_q\), \(\hat S_E\), and \(\hat c\) with the benchmark
metadata to identify whether an incorrect certificate first diverges in
query-scope extraction, evidence-coverage characterization, or the final
Boolean judgment. This analysis diagnoses errors in the model-reported
fields rather than  internal reasoning.

\section{Experiments}

Our experiments address five questions.
\begin{itemize}[leftmargin=9pt,itemsep=1pt,topsep=2pt]

\item \textbf{RQ1: Paired closure judgment.}
How reliably do LLMs distinguish
\textsc{Certified-Negative} from \textsc{Unknown} when only
query-relative coverage changes, with the question and observed facts
fixed?

\item \textbf{RQ2: Failure conditions.}
Are errors concentrated in partial-coverage or Scope-Mismatch cases,
and which coverage regimes yield the largest complete--partial gaps
across model families?

\item \textbf{RQ3: Prompting effects.}
Do explicit rules, chain-of-thought, abstention-aware prompting,
self-checking, and certificate elicitation improve paired discrimination,
or merely shift errors between over- and under-closure?

\item \textbf{RQ4: Diagnostic decomposition.}
At which certificate field does an incorrect output first differ from
the benchmark annotation: query scope, evidence coverage, or the final
coverage judgment?

\item \textbf{RQ5: Real-document transfer.}
Which CROWN-Synth failure patterns, particularly partial-coverage
over-closure, persist on CROWN-Real, and how do they vary across models,
prompts, and source structures?

\end{itemize}

\begin{table*}[t]
\centering
\scriptsize
\setlength{\tabcolsep}{2.0pt}
\renewcommand{\arraystretch}{0.70}
\begin{tabular*}{\textwidth}{@{\extracolsep{\fill}}l cccc cccc cccc}
\toprule
& \multicolumn{4}{c}{\textbf{Qwen3.5-9B}}
& \multicolumn{4}{c}{\textbf{Claude Haiku 4.5}}
& \multicolumn{4}{c}{\textbf{Gemma-4-12B}} \\
\cmidrule(lr){2-5}\cmidrule(lr){6-9}\cmidrule(lr){10-13}
\textbf{Condition}
& Acc$\uparrow$ & OCR$\downarrow$ & UCR$\downarrow$ & CS$\uparrow$
& Acc$\uparrow$ & OCR$\downarrow$ & UCR$\downarrow$ & CS$\uparrow$
& Acc$\uparrow$ & OCR$\downarrow$ & UCR$\downarrow$ & CS$\uparrow$ \\
\midrule
Naive & 70.6 & 37.2 & 21.6 & 42.8 & 73.7 & 40.8 & 11.8 & 47.6 & 61.5 & 76.0 & 0.9 & 23.2 \\
Def. & 79.9 & 26.9 & 13.2 & 60.0 & 83.5 & 13.3 & 19.6 & 67.1 & 75.0 & 44.6 & 5.4 & 50.5 \\
Def.+CoT & 83.9 & 27.9 & 4.4 & 68.1 & 89.9 & 12.4 & 7.8 & 79.8 & 85.2 & 24.8 & 4.9 & 70.9 \\
Def.+Abstain & 78.2 & 13.0 & 30.6 & 56.6 & 83.1 & 12.1 & 21.7 & 66.2 & 81.1 & 31.0 & 6.8 & 62.3 \\
Def.+Self-check & 74.5 & 8.8 & 42.3 & 49.4 & 85.2 & 16.6 & 13.1 & 70.4 & 73.1 & 51.8 & 2.0 & 46.6 \\
Cert. & 82.8 & 20.4 & 14.0 & 66.1 & 87.9 & 13.3 & 11.0 & 75.8 & 71.2 & 55.6 & 1.9 & 42.6 \\
Cert.+CoT & 85.7 & 24.6 & 4.0 & 71.9 & 87.9 & 17.6 & 6.6 & 75.9 & 79.1 & 40.3 & 1.6 & 58.3 \\
\bottomrule
\end{tabular*}
\caption{Overall results on CROWN-Synth (\%).}
\label{tab:main-results}
\end{table*}

\subsection{Models and Evaluation Conditions}

We evaluate three models from two open-weight families and one API-based
family: Qwen3.5-9B, Gemma-4-12B, and Claude Haiku 4.5. 

The evaluation conditions are summarized in Appendix~B. We evaluate seven LLM conditions.
Naive measures the model's default
treatment of missing support; Definition-aware states the task rule
explicitly. Its CoT, Abstain, and Self-check variants test step-by-step
reasoning, a conservative default to \textsc{Unknown}, and second-pass
revision. Certificate elicits
\((\hat S_q,\hat S_E,\hat c)\); only \(\hat c\) is mapped to the scored
label, while the two scope fields are retained for analysis.
Certificate+CoT adds reasoning before the same structured output.

\subsection{Experimental Protocol}

For each example, all LLM conditions receive the same question and
evidence context, use no few-shot demonstrations, and use greedy decoding
at temperature zero. Outputs are scored in the common two-label space.
Metric definitions use proportions in \([0,1]\); for readability, tables
report rates as percentages and rate differences in percentage points.
We report percentile 95\% confidence intervals from 10{,}000 bootstrap
replicates, resampling matched pairs for CROWN-Synth and A/B/C contrast sets for
CROWN-Real. Appendix~D reports the full condition
tables and bootstrap intervals.

\subsection{Overall Closure Profiles}

Table~\ref{tab:main-results} provides the aggregate view for RQ1 and
previews the prompting effects examined in RQ3. Under Naive prompting,
OCR exceeds UCR for all three models, revealing a default tendency to
license negative answers from non-closing evidence rather than remain
uncertain. Explicit task rules improve \(\mathrm{Acc}\) and CS for all three models,
and adding CoT further improves both metrics. However, these aggregate
gains do not consistently reduce OCR and UCR together: abstention lowers
OCR by increasing UCR, while self-checking and certificate elicitation have model-dependent
effects.

\noindent\textbf{Closure judgments remain unstable (RQ1).}
The evaluated LLMs exhibit some completeness-sensitive reasoning, but do
not reliably distinguish \textsc{Certified-Negative} from
\textsc{Unknown}. Explicit rules improve performance, yet the distinction
remains sensitive to the model and prompting condition. The central
limitation is therefore an unstable query-relative closure judgment.

\subsection{Failure Conditions}

\begin{table}[t]
\centering
\scriptsize
\setlength{\tabcolsep}{5.9pt}
\renewcommand{\arraystretch}{0.70}
\begin{tabular}{@{}ll ccc ccc@{}}
\toprule
&
& \multicolumn{3}{c}{\shortstack{\textbf{Pair: Complete}\\
                                  \textbf{vs.\ Partial}}}
& \multicolumn{3}{c}{\shortstack{\textbf{Pair: Complete}\\
                                  \textbf{vs.\ Scope-Mismatch}}} \\
\cmidrule(lr){3-5}\cmidrule(lr){6-8}
\textbf{Model} & \textbf{Condition}
& CS$\uparrow$
& \shortstack{Both-\\CN$\downarrow$}
& \shortstack{Both-\\UNK$\downarrow$}
& CS$\uparrow$
& \shortstack{Both-\\CN$\downarrow$}
& \shortstack{Both-\\UNK$\downarrow$} \\
\midrule
\multirow{4}{*}{Qwen}
& Naive     & 28.0 & 56.6 & 13.4 & 57.6 & 14.6 & 26.6 \\
& Def.      & 44.6 & 43.8 & 11.6 & 75.5 & 9.6  & 14.5 \\
& Def.+CoT  & 58.6 & 34.5 & 6.6  & 77.6 & 20.3 & 1.3  \\
& Cert.     & 59.3 & 30.2 & 10.0 & 72.9 & 9.8  & 17.1 \\
\midrule
\multirow{4}{*}{Haiku}
& Naive     & 26.1 & 67.9 & 6.0  & 69.0 & 13.3 & 17.3 \\
& Def.      & 66.2 & 18.6 & 15.0 & 67.9 & 7.9  & 24.2 \\
& Def.+CoT  & 72.6 & 20.5 & 7.0  & 87.0 & 4.3  & 8.6  \\
& Cert.     & 70.4 & 21.9 & 7.6  & 81.2 & 4.6  & 14.2 \\
\midrule
\multirow{4}{*}{Gemma}
& Naive     & 2.1  & 97.9 & 0.0  & 44.2 & 53.9 & 1.6 \\
& Def.      & 33.4 & 61.8 & 4.2  & 67.7 & 26.3 & 5.6 \\
& Def.+CoT  & 66.7 & 31.8 & 0.6  & 75.0 & 16.7 & 8.1 \\
& Cert.     & 17.0 & 82.7 & 0.2  & 68.2 & 28.2 & 3.4 \\
\bottomrule
\end{tabular}
\caption{Branch-specific CS and same-label pair outcomes on
CROWN-Synth (\%). The full outcome decomposition appears in
Table~\ref{tab:pair-full} in Appendix~D.}
\label{tab:pair-decomp}
\end{table}

Table~\ref{tab:pair-decomp} compares complete--partial pairs with
complete--Scope-Mismatch pairs. In both pair types, the complete
query-covering member has gold label \textsc{Certified-Negative}, and
the matched non-closing member has gold label \textsc{Unknown}. CS
requires both members to be correct. Both-CN indicates that both members
are predicted as \textsc{Certified-Negative}, producing over-closure on
the non-closing member. Both-UNK indicates that both members are predicted
as \textsc{Unknown}, producing under-closure on the complete member.
Across every model, CS is lower and Both-CN is higher
for complete--partial pairs than for complete--Scope-Mismatch pairs.
This indicates that, when coverage changes from complete to partial,
models often fail to switch their prediction from
\textsc{Certified-Negative} to \textsc{Unknown} and instead predict
\textsc{Certified-Negative} for both members.

Having established that complete--partial pairs produce more errors than
complete--Scope-Mismatch pairs, we next examine which coverage regimes
account for these failures. Table~\ref{tab:coverage-regimes} divides the
1{,}250 complete--partial pairs across L1--L4. The first four columns
report regime-specific CS, while the final two columns report the
separate correct rates for the complete and partial members within
L3 complete--partial pairs.

As shown in Table~\ref{tab:coverage-regimes}, L3 has the lowest or
tied-lowest CS in every model--condition row. Within L3, the correct rate
for the complete member ranges from 82.4 to 100.0\%, whereas that for the
partial member ranges from 0.0 to 27.9\%. Thus, the low L3 CS is driven
primarily by predicting \textsc{Certified-Negative} for the partial
member, rather than by errors on the complete member. This
complete--partial asymmetry holds across all four complete and four
partial source-type families in the domain-crossed L3 inventory
(Appendix~D, Table~\ref{tab:l3-family}).

\noindent\textbf{Errors concentrate in complete--partial pairs,
especially L3 (RQ2).}
Compared with Scope-Mismatch pairs, complete--partial pairs have lower
CS and more Both-CN outcomes across every model and condition shown.
Within the complete--partial branch, L3 has the lowest or tied-lowest CS:
models usually answer the implicit-complete member correctly but often
also answer the implicit-partial member as
\textsc{Certified-Negative}.

\subsection{Prompting Effects}

Table~\ref{tab:main-results} shows that explicit rules and CoT often
improve aggregate performance, but RQ3 asks whether these gains repair
the failure modes identified above or merely shift errors across items.
We therefore compare predictions item by item.
Table~\ref{tab:prompt-shifts} reports the accuracy change after adding
CoT within each item group. Positive values indicate more corrections
than regressions.

\begin{table}[t]
\centering
\scriptsize
\setlength{\tabcolsep}{5.1pt}
\renewcommand{\arraystretch}{0.70}
\begin{tabular}{@{}ll rrrr rr@{}}
\toprule
&
& \multicolumn{4}{c}{\shortstack{\textbf{Regime-Specific}\\
                                  \textbf{CS}}}
& \multicolumn{2}{c}{\shortstack{\textbf{L3 Member}\\
                                  \textbf{Correct Rate}}} \\
\cmidrule(lr){3-6}\cmidrule(lr){7-8}
\textbf{Model} & \textbf{Cond.}
& L1 & L2 & L3 & L4
& \shortstack{Comp.-\\CN$\uparrow$}
& \shortstack{Part.-\\UNK$\uparrow$} \\
\midrule
\multirow{4}{*}{Qwen}
& Naive    & 45.4 & 36.7 & 12.5 & 17.3 & 86.5  & 24.7 \\
& Def.     & 58.8 & 70.0 & 9.6  & 39.7 & 89.7  & 19.9 \\
& Def.+CoT & 94.9 & 97.1 & 8.7  & 33.3 & 91.7  & 16.0 \\
& Cert.    & 85.0 & 79.2 & 7.7  & 65.1 & 86.9  & 18.6 \\
\midrule
\multirow{4}{*}{Haiku}
& Naive    & 38.0  & 30.4 & 9.3  & 26.6 & 96.2 & 13.1 \\
& Def.     & 97.8  & 99.4 & 10.6 & 57.1 & 82.4 & 27.9 \\
& Def.+CoT & 100.0 & 99.4 & 15.1 & 75.6 & 92.9 & 22.1 \\
& Cert.    & 94.6  & 99.7 & 17.6 & 69.6 & 99.4 & 18.3 \\
\midrule
\multirow{4}{*}{Gemma}
& Naive    & 3.8  & 0.0  & 0.0 & 4.5  & 100.0 & 0.0 \\
& Def.     & 49.8 & 42.8 & 0.0 & 40.7 & 100.0 & 0.0 \\
& Def.+CoT & 99.4 & 99.4 & 6.4 & 61.5 & 98.7  & 6.7 \\
& Cert.    & 22.4 & 16.9 & 0.0 & 28.8 & 100.0 & 0.0 \\
\bottomrule
\end{tabular}
\caption{Regime-specific CS on 1{,}250 complete--partial pairs
(\%). L1--L4 report CS within each coverage regime. Comp.-CN and
Part.-UNK report the correct rates for the complete and partial members
of L3, respectively. Full results appear in
Table~\ref{tab:full-partial-regimes} in Appendix~D.}
\label{tab:coverage-regimes}
\end{table}

\begin{table}[t]
\centering
\scriptsize
\setlength{\tabcolsep}{6.5pt}
\renewcommand{\arraystretch}{0.70}
\begin{tabular}{@{}llrrrr@{}}
\toprule
\textbf{Model} & \textbf{Transition}
& \textbf{Complete} & \textbf{L3-Part.} & \textbf{L4-Part.} & \textbf{SM} \\
\midrule
\multirow{2}{*}{Qwen}
& Def.\(\rightarrow\)Def.+CoT
& +8.9 & -3.8 & -7.7 & -11.1 \\
& Cert.\(\rightarrow\)Cert.+CoT
& +9.9 & -4.8 & -29.8 & -5.7 \\
\midrule
\multirow{2}{*}{Haiku}
& Def.\(\rightarrow\)Def.+CoT
& +11.8 & -5.8 & -4.2 & +3.6 \\
& Cert.\(\rightarrow\)Cert.+CoT
& +4.4 & +1.3 & -37.2 & -0.6 \\
\midrule
\multirow{2}{*}{Gemma}
& Def.\(\rightarrow\)Def.+CoT
& +0.5 & +6.7 & +6.1 & +9.8 \\
& Cert.\(\rightarrow\)Cert.+CoT
& +0.3 & +0.6 & +19.9 & +9.9 \\
\bottomrule
\end{tabular}
\caption{Item-level accuracy change after adding CoT
(percentage points). L3-Part.\ and L4-Part.\ denote partial members in
L3 and L4, respectively; SM denotes Scope-Mismatch members. Positive
values indicate accuracy gains, and negative values indicate losses.
Full transition results appear in Table~\ref{tab:all-transitions} in
Appendix~D.}
\label{tab:prompt-shifts}
\end{table}

CoT does not provide a uniform correction. Under Definition-aware
prompting, it repairs complete cases for Qwen and Haiku but regresses on
L3 and L4 partial evidence. With certificates, the regression
concentrates on L4 for both models, while the L3 change is small and
differs in sign. Gemma instead shows broader repairs on non-closing
cases. Thus, the same instruction can move models in opposite directions.

As shown in Table~\ref{tab:all-transitions} in
Appendix~D, abstention-aware prompting produces a more predictable shift toward \textsc{Unknown}: it repairs many non-closing cases but creates new
under-closure on complete evidence. Self-checking is less consistent and
does not yield a common improvement pattern across models.

\noindent\textbf{Prompting redistributes rather than removes closure
errors (RQ3).}
Explicit rules and reasoning can improve aggregate performance, but no
intervention consistently repairs partial-coverage over-closure while
preserving correct judgments on complete evidence across models.
Prompting often redistributes errors between over- and under-closure
and does not consistently improve discrimination between complete
query-covering evidence and non-closing evidence.

\subsection{Diagnostic Decomposition}

To answer RQ4, we analyze the three fields of each incorrect
completeness certificate
\(C=(\hat S_q,\hat S_E,\hat c)\).
We compare the reported query scope \(\hat S_q\) with \(S(q)\), the
reported evidence coverage \(\hat S_E\) with the gold evidence scope
\(S(E)\) and its completeness status, and the final judgment
\(\hat c\) with \(c\). Each error is assigned to its earliest erroneous
field: query-scope when \(\hat S_q\) is incorrect, evidence-coverage when
\(\hat S_E\) is incorrect after an adequate query scope, and Boolean when
both scope fields are adequate but \(\hat c\neq c\).

\begin{table}[t]
\centering
\scriptsize
\setlength{\tabcolsep}{11.2pt}
\renewcommand{\arraystretch}{0.70}
\begin{tabular}{@{}lrrrr@{}}
\toprule
\textbf{Model} & \textbf{\#Err.}
& \shortstack{\textbf{Query} $\hat S_q$}
& \shortstack{\textbf{Evidence} $\hat S_E$}
& \shortstack{\textbf{Boolean} $\hat c$} \\
\midrule
Qwen  & 858  & 24.4 & 50.1 & 25.5 \\
Haiku & 606  & 23.9 & 45.4 & 30.7 \\
Gemma & 1438 & 26.7 & 72.0 & 1.3  \\
\bottomrule
\end{tabular}
\caption{Earliest erroneous field among incorrect Certificate outputs.
\#Err.\ is the error count; the remaining columns report percentages.}
\label{tab:certificate-errors}
\end{table}

Evidence-coverage errors in \(\hat S_E\) form the largest category for
all three models and are especially concentrated for Gemma. Qwen and
Haiku show more distributed profiles, including more errors in the final
judgment \(\hat c\). Thus, the earliest reported error most often lies in the characterization
of evidence scope or completeness, rather than only in the final Boolean
judgment.

\noindent\textbf{Evidence coverage is the main reported failure point
(RQ4).}
The certificate fields show that the earliest reported error is most
often in \(\hat S_E\), where models mischaracterize the evidence scope
or completeness rather than only erring in the final coverage judgment.

\begin{table}[t]
\centering
\scriptsize
\setlength{\tabcolsep}{3.0pt}
\renewcommand{\arraystretch}{0.70}
\begin{tabular}{@{}ll rrrr rr@{}}
\toprule
&
& \multicolumn{4}{c}{\textbf{Proceedings}}
& \multicolumn{2}{c}{\textbf{Medical}} \\
\cmidrule(lr){3-6}\cmidrule(lr){7-8}
\textbf{Model} & \textbf{Cond.}
& A-CN$\uparrow$ & B-UNK$\uparrow$ & C-UNK$\uparrow$
& \(\Delta_{B-C}\)
& A-CN$\uparrow$ & B-UNK$\uparrow$ \\
\midrule

\multirow{4}{*}{Qwen}
& Naive    & 35.0 & 93.6 & 73.7 & 19.9 & 24.7 & 100.0 \\
& Def.     & 0.0  & 100.0 & 100.0 & 0.0 & 8.6 & 100.0 \\
& Def.+CoT & 88.7 & 99.6 & 66.5 & 33.1 & 94.8 & 100.0 \\
& Cert.    & 93.2 & 100.0 & 91.7 & 8.3 & 71.5 & 98.5 \\
\midrule

\multirow{4}{*}{Haiku}
& Naive    & 62.8 & 63.2 & 20.3 & 42.9 & 86.5 & 41.2 \\
& Def.     & 99.2 & 32.7 & 3.0  & 29.7 & 99.3 & 24.3 \\
& Def.+CoT & 100.0 & 98.1 & 27.8 & 70.3 & 100.0 & 100.0 \\
& Cert.    & 95.5 & 100.0 & 12.0 & 88.0 & 91.4 & 100.0 \\
\midrule

\multirow{4}{*}{Gemma}
& Naive    & 97.0 & 1.1  & 0.8 & 0.4 & 99.3 & 10.5 \\
& Def.     & 99.6 & 0.0  & 0.0 & 0.0 & 100.0 & 41.6 \\
& Def.+CoT & 100.0 & 93.6 & 3.4 & 90.2 & 98.1 & 100.0 \\
& Cert.    & 100.0 & 98.9 & 1.9 & 97.0 & 97.8 & 100.0 \\
\bottomrule
\end{tabular}

\caption{CROWN-Real variant-level correct rates (\%) for the four core
LLM conditions. Full results are reported in
Table~\ref{tab:real-transfer-full} in Appendix~D.}
\label{tab:real-transfer}
\end{table}

\subsection{Transfer to CROWN-Real}

Table~\ref{tab:real-transfer} reports how the models handle the A/B/C
variants on two real-document sources. A-CN is the percentage of
query-covering A variants predicted as
\textsc{Certified-Negative}; B-UNK and C-UNK are the percentages of B
and C variants predicted as \textsc{Unknown}, respectively. For
Proceedings,
\(\Delta_{B-C}=\text{B-UNK}-\text{C-UNK}\); a positive value means that
C is more error-prone than B. Medical C is an explicit-partial control
and is excluded from the implicit-partial comparison.

The results vary  across models, prompts, and document
sources. One pattern, however, is consistent. In Proceedings, B is
complete for one queried collection but does not cover the full query
scope, corresponding to Scope-Mismatch in CROWN-Synth. C contains only a
subset of titles from the queried collections, corresponding to partial
coverage. Across all 15 model--condition cells, \(\Delta_{B-C}\) is nonnegative,
and its bootstrap interval excludes zero in 12
(Appendix~D, Table~\ref{tab:bootstrap-real}).
Because B and C are both labeled \textsc{Unknown}, the lower C-UNK
rates show that models more often predict \textsc{Certified-Negative}
for partial C than for narrower-complete B. This mirrors the ordering
in CROWN-Synth, where complete--partial pairs have lower CS and higher
Both-CN rates than complete--Scope-Mismatch pairs.

\noindent\textbf{Partial evidence remains at least as difficult as
Scope-Mismatch evidence (RQ5).}
In CROWN-Synth and the Proceedings component of CROWN-Real, partial
evidence is at least as error-prone as evidence that is complete only for
a narrower scope. However, accuracy on query-covering evidence and the
size of the partial-coverage error differ across models, prompting
conditions, and document sources.

\section{Conclusion}

We introduced completeness-sensitive negative reasoning and CROWN-QA,
combining a controlled paired core with a real-document contrast-set
evaluation. LLMs show partial but unstable ability
to distinguish \textsc{Certified-Negative} from \textsc{Unknown}.
CROWN-Synth exposes a pronounced asymmetry in the implicit source-type
regime: models usually classify complete framings correctly, but often
also label the matched partial framings as
\textsc{Certified-Negative} instead of \textsc{Unknown}.
Prompting redistributes errors between over- and under-closure without a
consistent remedy, while certificate analysis shows that the earliest
reported error most often lies in evidence-coverage characterization.
On CROWN-Real Proceedings, partial evidence is at least as error-prone
as narrower-scope complete evidence across every model and condition,
although accuracy and error balance vary. Taken together, these results
isolate a challenge beyond detecting absent support: determining whether
the available evidence completely covers the query scope.

\bibliography{crown_qa_references}

\clearpage
\onecolumn
\appendix

\section{Appendix A. Dataset Examples}
\label{app:data}

\subsection{Dataset Format and Worked Examples}
\label{app:data-format}

CROWN-Synth and CROWN-Real are stored as JSONL files, with one example
per line. The examples below show shortened versions of the text given
to the models. Full records and metadata are included in the submitted
files.

\paragraph{CROWN-Synth worked pair.}

The following example is the L3 pair
\texttt{proceedings\_0252}. MLConf and the listed paper titles are
synthetic. Both members use the same question and the same three listed
titles. Only the source-type phrase in the first sentence changes.

\noindent Question:
Did MLConf 2025 publish a paper whose title contains
``structured prompt plans''?

\noindent Shared fictional paper titles:
\begin{itemize}
    \item Reliable Evaluation for Planning Agents
    \item Incremental Updates for Knowledge Stores
    \item Scoring Citations in Literature Reviews
\end{itemize}

No listed title contains ``structured prompt plans.''

\begin{table}[H]
\centering
\small
\setlength{\tabcolsep}{6pt}
\renewcommand{\arraystretch}{1.05}
\begin{tabular}{p{0.16\textwidth} p{0.38\textwidth}
                p{0.38\textwidth}}
\toprule
& \textbf{Complete member} & \textbf{Partial member} \\
\midrule

Coverage sentence
&
The master index for the MLConf 2025 proceedings lists these paper
titles.
&
The highlights digest for the MLConf 2025 proceedings lists these paper
titles.
\\

Gold label
&
\textsc{Certified-Negative}
&
\textsc{Unknown}
\\

\bottomrule
\end{tabular}
\caption{Shortened CROWN-Synth L3 pair. The question and listed titles
are identical; only the source-type phrase changes.}
\label{tab:synth-worked-pair}
\end{table}

\paragraph{CROWN-Real worked contrast set.}

The following medical contrast set uses the same question and target
phrase in all three variants. The target phrase is absent from every
displayed context.

\noindent Question:
For Atropine label version 2026-03-10, does Section 5, Warnings and
Precautions, of the Full Prescribing Information contain the exact
phrase ``Impaired Renal Function''?

\begin{table}[H]
\centering
\small
\setlength{\tabcolsep}{5pt}
\renewcommand{\arraystretch}{1.05}
\begin{tabular}{p{0.14\textwidth} p{0.27\textwidth}
                p{0.27\textwidth} p{0.27\textwidth}}
\toprule
& \textbf{Variant A} & \textbf{Variant B} & \textbf{Variant C} \\
\midrule

Evidence shown
&
Full Section 5, beginning with ``5. WARNINGS AND PRECAUTIONS'' and
ending at ``6. ADVERSE REACTIONS.''
&
Subsection 5.2, ``Elevation of Blood Pressure,'' ending at the heading
for subsection 5.3.
&
``HIGHLIGHTS OF PRESCRIBING INFORMATION,'' including the statement that
the Highlights do not contain all information needed to use the drug.
\\

Coverage
&
Complete for the full queried Section 5
&
Complete only for subsection 5.2
&
Non-exhaustive Highlights
\\

Gold label
&
\textsc{Certified-Negative}
&
\textsc{Unknown}
&
\textsc{Unknown}
\\

\bottomrule
\end{tabular}
\caption{Shortened CROWN-Real medical contrast set. The question and
target phrase are fixed, while the displayed document range changes.}
\label{tab:real-worked-set}
\end{table}

\section{Appendix B. Prompt Templates}
\label{app:prompts}

Table~\ref{tab:conditions} summarizes the seven LLM conditions, the
number of calls required by each condition, and the intervention being
tested. The conditions range from direct-label prediction to reasoning,
abstention-aware prompting, self-revision, and structured
scope-and-coverage elicitation. We next provide the exact prompt
templates and shared components used across conditions.

\begin{center}
\footnotesize
\setlength{\tabcolsep}{2.5pt}
\renewcommand{\arraystretch}{1.05}
\begin{tabular}{l c p{0.47\columnwidth}}
\toprule
\textbf{Condition} & \textbf{Calls} & \textbf{Intervention} \\
\midrule
Naive & 1 & Direct label without the task rule. \\
Definition-aware & 1 & Direct label with the task rule. \\
Definition-aware+CoT & 1 & Adds step-by-step reasoning. \\
Definition-aware+Abstain & 1 & Adds a conservative default to
\textsc{Unknown}. \\
Definition-aware+Self-check & 2 & Reviews and may revise an initial
label. \\
Certificate & 1 & Elicits $(\hat S_q,\hat S_E,\hat c)$. \\
Certificate+CoT & 1 & Adds reasoning to certificate elicitation. \\
\bottomrule
\end{tabular}
\captionof{table}{Summary of the evaluation conditions and their
corresponding interventions.}
\label{tab:conditions}
\end{center}

All LLM conditions receive the same base question and evidence context,
and no few-shot demonstrations are used. For the API model, the common
system message is supplied through the system role. For open-weight
models, it is placed in the system segment of the corresponding chat
template.

The prompt templates below use bracketed block names such as
[Shared Task Rule] and [Input Block]. At runtime, each bracketed name is
replaced by the complete text of the corresponding common component
defined below. The bracketed names themselves are not sent to the model.
The order shown inside each prompt box is the exact order used at
runtime, with one blank line between consecutive blocks.

\subsection{Common Components}

\subsubsection{System Message}

The following system message is used for every LLM call.

\begin{tcolorbox}[colback=gray!5, colframe=gray!80,
sharp corners, boxrule=0.5pt]
You are an evidence-grounded question-answering classifier. Use only the
provided evidence and do not use external knowledge. Follow the requested
output format exactly and do not add unrequested text.
\end{tcolorbox}

\subsubsection{Shared Task Rule}

The following rule block is used by Definition-aware and its CoT,
Abstain, and Self-check variants. It is omitted from Naive, Certificate,
and Certificate+CoT.

\begin{tcolorbox}[colback=gray!5, colframe=gray!80,
sharp corners, boxrule=0.5pt]
Apply the following rules:

The queried fact is not supported by the observed factual content. Your
task is to decide whether this observed absence is licensed as a negative
answer.

Use CERTIFIED\_NEGATIVE only when the evidence is complete for the exact scope
        of the question.

Use UNKNOWN when the evidence is partial, sampled, incomplete,
unspecified, ambiguous, or complete only for a different scope.

Evidence that is complete only for a different entity, time period, attribute,
        or collection is not complete for the question.
\end{tcolorbox}

\subsubsection{Direct-Label Output Block}

The following output block is used when the LLM directly returns one of
the two task labels.

\begin{tcolorbox}[colback=gray!5, colframe=gray!80,
sharp corners, boxrule=0.5pt]
Return exactly one of the following labels and nothing else:

CERTIFIED\_NEGATIVE

UNKNOWN
\end{tcolorbox}

\subsubsection{Input Block}

The following input block is appended to every LLM user message.

\begin{tcolorbox}[colback=gray!5, colframe=gray!80,
sharp corners, boxrule=0.5pt]
Evidence:

\{context\}

Question:

\{question\}
\end{tcolorbox}

Direct-label outputs are parsed according to the condition-specific
format given below. The machine-readable labels CERTIFIED\_NEGATIVE and
UNKNOWN correspond respectively to \textsc{Certified-Negative} and
\textsc{Unknown}. Output normalization is specified in Appendix~C.

\subsection{Naive Prompt}

The Naive condition omits the shared task rule and measures the model's
default treatment of missing evidence. The following box shows the
complete user-message assembly.

\begin{tcolorbox}[colback=gray!5, colframe=gray!80,
sharp corners, boxrule=0.5pt]
Classify whether the observed absence licenses a negative answer using
only the provided evidence.

[Direct-Label Output Block]

[Input Block]
\end{tcolorbox}

Thus, the Naive condition consists of the common system message and one
user message containing the three components above.

\subsection{Definition-Aware Prompt}

The Definition-aware condition provides the task rule but does not
request intermediate scope or coverage judgments.

\begin{tcolorbox}[colback=gray!5, colframe=gray!80,
sharp corners, boxrule=0.5pt]
Classify whether the observed absence licenses a negative answer using only the provided evidence.

[Shared Task Rule]

[Direct-Label Output Block]

[Input Block]
\end{tcolorbox}

The LLM directly generates the final label in one call.

\subsection{Definition-Aware+CoT Prompt}

The Definition-aware+CoT condition adds a general step-by-step reasoning
instruction to the Definition-aware condition. The LLM still generates
the final label directly; no structured certificate or fixed decision
rule is used.

\begin{tcolorbox}[colback=gray!5, colframe=gray!80,
sharp corners, boxrule=0.5pt]
Classify whether the observed absence licenses a negative answer using only the provided evidence.

[Shared Task Rule]

Reason step by step before choosing the final label.

End the response with exactly one of the following lines:

FINAL LABEL: CERTIFIED\_NEGATIVE

FINAL LABEL: UNKNOWN

[Input Block]
\end{tcolorbox}

Only the final line beginning with ``FINAL LABEL:'' is used for scoring.

\subsection{Definition-Aware+Abstain Prompt}

The Definition-aware+Abstain condition uses the shared task rule and adds
a conservative default for uncertain cases.

\begin{tcolorbox}[colback=gray!5, colframe=gray!80,
sharp corners, boxrule=0.5pt]
Classify whether the observed absence licenses a negative answer using only the provided evidence.

[Shared Task Rule]

Adopt a conservative policy. If it is uncertain whether the evidence is complete
        for the exact scope of the question, return UNKNOWN.

[Direct-Label Output Block]

[Input Block]
\end{tcolorbox}

This condition tests whether over-closure can be reduced by returning
\textsc{Unknown} more readily and whether this increases under-closure.

\subsection{Definition-Aware+Self-Check Prompt}

Definition-aware+Self-check uses two LLM calls. The first call uses the
complete Definition-aware prompt above. Its normalized output is inserted
as \{initial\_label\} in the second-call prompt below.

\begin{tcolorbox}[colback=gray!5, colframe=gray!80,
sharp corners, boxrule=0.5pt]
Review the following initial label and revise it if necessary:

\{initial\_label\}

[Shared Task Rule]

Do not assume that the initial label is correct. Verify whether the evidence is complete for the exact scope of the question. Return the label implied by the task rule.

[Direct-Label Output Block]

[Input Block]
\end{tcolorbox}

Only the second-call response is used as the final prediction. 

\subsection{Certificate Prompt}

The Certificate condition uses one structured LLM call. Unlike the
direct-label conditions, it does not include the Shared Task Rule,
because the LLM is not asked to choose between the two labels. Instead,
the model returns the query scope, evidence coverage scope, and coverage
judgment. A fixed decision rule then maps the
\texttt{complete\_for\_query} field to the final prediction.

\begin{tcolorbox}[colback=gray!5, colframe=gray!80,
sharp corners, boxrule=0.5pt]
Do not output a final answer label.

Using only the provided evidence, return a valid JSON object with exactly
the following three keys:

\{

"query\_scope": "...",

"evidence\_coverage\_scope": "...",

"complete\_for\_query": true or false

\}

Use the fields as follows:

"query\_scope" must describe all constraints required by the question,
such as the relevant entity, time period, attribute, and collection.

"evidence\_coverage\_scope" must describe what the evidence covers and
whether that coverage is complete, partial, or unspecified.

Set "complete\_for\_query" to true only when the evidence establishes
complete coverage of the entire query scope. Evidence that is complete
only for a different scope is not complete for the question. Set the field to false when coverage is partial, sampled, unspecified, ambiguous, or mismatched with the query scope.

Return only the JSON object. Do not include an explanation, markdown
formatting, a final label, or additional keys.

[Input Block]
\end{tcolorbox}

\subsection{Certificate+CoT Prompt}

The Certificate+CoT condition uses one structured LLM call. It asks the
model to reason about the query scope, evidence coverage scope, and
coverage relation before returning the same three certificate fields as
Certificate. It does not include the Shared Task Rule or the Direct-Label
Output Block.

\begin{tcolorbox}[colback=gray!5, colframe=gray!80,
sharp corners, boxrule=0.5pt]
Do not output a final answer label.

First reason step by step about the following three points:

1. What is the exact query scope?

2. What scope does the evidence claim to cover?

3. Does the evidence coverage close the entire query scope?

Then return a final JSON object.

Use only the provided evidence. Do not use external knowledge.

The final JSON object must have exactly the following three keys:

\{

"query\_scope": "...",

"evidence\_coverage\_scope": "...",

"complete\_for\_query": true or false

\}

Use the fields as follows:

"query\_scope" must describe all constraints required by the question,
such as the relevant entity, time period, attribute, and collection.

"evidence\_coverage\_scope" must describe what the evidence covers and
whether that coverage is complete, partial, or unspecified.

Set "complete\_for\_query" to true only when the evidence establishes
complete coverage of the entire query scope. Evidence that is complete
only for a different scope is not complete for the question. Set the field to false when coverage is partial, sampled, unspecified, ambiguous, or mismatched with the query scope.

Return the response in exactly this format:

REASONING:

<brief step-by-step reasoning>

FINAL\_JSON:

\{

"query\_scope": "...",

"evidence\_coverage\_scope": "...",

"complete\_for\_query": true or false

\}

Do not wrap the JSON in markdown code fences. Do not include a final label, additional keys, or any text after the JSON object. The value of "complete\_for\_query" must be a JSON boolean, not a string.

[Input Block]
\end{tcolorbox}

Only the JSON object following ``FINAL\_JSON:'' is parsed for scoring.
Its Boolean field is mapped to the two-label space using the same fixed
mapping as Certificate. The reasoning text and the two scope fields are
retained in the experiment logs for diagnostic analysis.

\section{Appendix C. Implementation Details}
\label{app:implementation}

\subsection{Models and Runtime Environment}

We evaluate three instruction-tuned LLMs spanning open-weight and API
access: Qwen3.5-9B, Gemma-4-12B, and Claude Haiku 4.5. For each
benchmark, all models are evaluated on the same frozen dataset version
and receive no in-context examples.

The open-weight models were run on one NVIDIA A100 GPU in Google Colab.
Claude Haiku 4.5 was accessed through the Anthropic API. Definition-aware+Self-check normally used two sequential calls per
example; all other conditions normally used one call. If a response
could not be parsed, the same call was repeated once with the same input,
prompt, and generation settings. No further retries were made. All calls
used temperature \(0\), and the open-weight models used greedy decoding. The submitted artifact contains
the datasets, exact prompts, raw outputs, inference and scoring code,
model configuration files, and the Python environment used for the
reported experiments.

\subsection{Generation Limit}

All calls used a maximum output length of 5120 new tokens. No input
question or evidence context was truncated.

\subsection{Output Parsing and Scoring}

All predictions are scored in the two-label space
\(\{\textsc{Certified-Negative},\textsc{Unknown}\}\). The parser follows
the output format specified for each condition in Appendix~B.
For Naive, Definition-aware, and Definition-aware+Abstain, the parser
reads the direct label. For Self-check, the first response provides the
initial label, and the second response is used as the final prediction
for scoring. For Definition-aware+CoT, the parser reads the final line
beginning with \texttt{FINAL LABEL:}.

For Certificate, the parser reads the returned JSON object. For
Certificate+CoT, it reads the JSON object following
\texttt{FINAL\_JSON:}. The parsed object must contain
\texttt{query\_scope}, \texttt{evidence\_coverage\_scope}, and
\texttt{complete\_for\_query}. The Boolean field is mapped as follows:
\[
\texttt{true}\mapsto\textsc{Certified-Negative},
\qquad
\texttt{false}\mapsto\textsc{Unknown}.
\]

If the first response could not be parsed, the same call was repeated
once with the same input, prompt, and generation settings. When the
second response was parseable, it was used for scoring. If the second
response also could not be parsed, the example was counted as incorrect
and remained in the denominator; it was never mapped to
\textsc{Unknown}. The ``Off'' column of
Table~\ref{tab:pair-full} reports the rate of examples that remained
unparseable after this retry.

For the Certificate condition in
Table~\ref{tab:certificate-errors}, each incorrect output is assigned to
the first field that fails the rule-based check. The script first
compares \texttt{query\_scope} with the gold query scope. If that check
passes, it compares \texttt{evidence\_coverage\_scope} with the gold
evidence scope and coverage status. If both scope fields pass but
\texttt{complete\_for\_query} is incorrect, the error is assigned to the
Boolean field.

\section{Appendix D. Additional Experimental Results and Analyses}
\label{app:extended-results}

\subsection{Full Pair Outcome Decomposition}
Table~\ref{tab:pair-full} extends the pair-level analysis in the main
text to all evaluation conditions. Across all evaluation conditions,
complete--partial pairs produce Both-CN more often than
complete--Scope-Mismatch pairs. This indicates that models are more
prone to over-close on partial evidence than on Scope-Mismatch evidence.
Reversed outcomes and output-format failures are negligible, so most
pair failures arise from assigning the same label to both members.

\begin{table}[H]
\centering
\scriptsize
\setlength{\tabcolsep}{5.0pt}
\renewcommand{\arraystretch}{0.9}
\begin{tabular}{@{}ll ccccc ccccc@{}}
\toprule
&
& \multicolumn{5}{c}{\textbf{Pair: Complete vs.\ Partial}}
& \multicolumn{5}{c}{\textbf{Pair: Complete vs.\ Scope-Mismatch}} \\
\cmidrule(lr){3-7}\cmidrule(lr){8-12}
\textbf{Model} & \textbf{Condition}
& CS$\uparrow$ & Both-CN$\downarrow$ & Both-UNK$\downarrow$
& Rev.$\downarrow$ & Off$\downarrow$
& CS$\uparrow$ & Both-CN$\downarrow$ & Both-UNK$\downarrow$
& Rev.$\downarrow$ & Off$\downarrow$ \\
\midrule
\multirow{7}{*}{\textbf{Qwen}}
& Naive
& 28.0 & 56.6 & 13.4 & 2.0 & 0.0
& 57.6 & 14.6 & 26.6 & 1.1 & 0.0 \\
& Def.
& 44.6 & 43.8 & 11.6 & 0.0 & 0.0
& 75.5 & 9.6 & 14.5 & 0.4 & 0.0 \\
& Def.+CoT
& 58.6 & 34.5 & 6.6 & 0.2 & 0.1
& 77.6 & 20.3 & 1.3 & 0.5 & 0.3 \\
& Def.+Abstain
& 53.7 & 22.1 & 23.9 & 0.3 & 0.0
& 59.5 & 3.4 & 36.9 & 0.2 & 0.0 \\
& Def.+Self-check
& 48.6 & 14.3 & 36.4 & 0.7 & 0.0
& 50.2 & 2.2 & 47.3 & 0.2 & 0.0 \\
& Cert.
& 59.3 & 30.2 & 10.0 & 0.6 & 0.0
& 72.9 & 9.8 & 17.1 & 0.2 & 0.0 \\
& Cert.+CoT
& 65.2 & 32.3 & 1.1 & 0.2 & 1.1
& 78.6 & 14.4 & 5.0 & 0.7 & 1.4 \\
\midrule
\multirow{7}{*}{\textbf{Haiku}}
& Naive
& 26.1 & 67.9 & 6.0 & 0.0 & 0.0
& 69.0 & 13.3 & 17.3 & 0.4 & 0.0 \\
& Def.
& 66.2 & 18.6 & 15.0 & 0.1 & 0.0
& 67.9 & 7.9 & 24.2 & 0.0 & 0.0 \\
& Def.+CoT
& 72.6 & 20.5 & 7.0 & 0.0 & 0.0
& 87.0 & 4.3 & 8.6 & 0.0 & 0.0 \\
& Def.+Abstain
& 67.8 & 15.7 & 16.6 & 0.0 & 0.0
& 64.7 & 8.4 & 26.7 & 0.2 & 0.0 \\
& Def.+Self-check
& 71.2 & 19.9 & 8.9 & 0.0 & 0.0
& 69.6 & 13.1 & 17.1 & 0.2 & 0.0 \\
& Cert.
& 70.4 & 21.9 & 7.6 & 0.1 & 0.0
& 81.2 & 4.6 & 14.2 & 0.0 & 0.0 \\
& Cert.+CoT
& 64.2 & 29.9 & 5.8 & 0.0 & 0.0
& 87.6 & 5.0 & 7.2 & 0.2 & 0.0 \\
\midrule
\multirow{7}{*}{\textbf{Gemma}}
& Naive
& 2.1 & 97.9 & 0.0 & 0.0 & 0.0
& 44.2 & 53.9 & 1.6 & 0.2 & 0.0 \\
& Def.
& 33.4 & 61.8 & 4.2 & 0.6 & 0.0
& 67.7 & 26.3 & 5.6 & 0.4 & 0.0 \\
& Def.+CoT
& 66.7 & 31.8 & 0.6 & 0.9 & 0.0
& 75.0 & 16.7 & 8.1 & 0.2 & 0.0 \\
& Def.+Abstain
& 55.3 & 39.4 & 5.3 & 0.0 & 0.0
& 69.4 & 22.2 & 8.1 & 0.3 & 0.0 \\
& Def.+Self-check
& 34.5 & 65.3 & 0.0 & 0.2 & 0.0
& 58.7 & 37.5 & 3.1 & 0.6 & 0.0 \\
& Cert.
& 17.0 & 82.7 & 0.2 & 0.1 & 0.0
& 68.2 & 28.2 & 3.4 & 0.2 & 0.0 \\
& Cert.+CoT
& 36.4 & 61.9 & 1.5 & 0.2 & 0.0
& 80.2 & 18.3 & 1.4 & 0.2 & 0.0 \\
\bottomrule
\end{tabular}
\caption{Full pair-level outcome decomposition across models and
conditions (\%). CS denotes the correct paired outcome
\((\textsc{Certified-Negative},\textsc{Unknown})\). Both-CN and
Both-UNK assign the same label to both members; Rev.\ denotes the
reversed \((\textsc{Unknown},\textsc{Certified-Negative})\) outcome,
and Off denotes output-format failures.}
\label{tab:pair-full}
\end{table}

The full-condition decomposition therefore reinforces the RQ2 finding
that partial evidence is a more persistent source of over-closure than
Scope-Mismatch evidence.

\subsection{Full Complete--Partial Regime Analysis}

\begin{table}[H]
\centering
\scriptsize
\setlength{\tabcolsep}{10.0pt}
\renewcommand{\arraystretch}{0.9}
\begin{tabular}{@{}ll rrrr rr@{}}
\toprule
&
& \multicolumn{4}{c}{\shortstack{\textbf{Regime-Specific} \textbf{CS}}}
& \multicolumn{2}{c}{\shortstack{\textbf{L3 Member} \textbf{Correct Rate}}} \\
\cmidrule(lr){3-6}\cmidrule(lr){7-8}
\textbf{Model} & \textbf{Cond.}
& L1 & L2 & L3 & L4
& \shortstack{Comp.-CN$\uparrow$}
& \shortstack{Part.-UNK$\uparrow$} \\
\midrule
\multirow{7}{*}{{Qwen}}
& Naive           & 45.4  & 36.7  & 12.5  & 17.3 & 86.5  & 24.7  \\
& Def.            & 58.8  & 70.0  & 9.6   & 39.7 & 89.7  & 19.9  \\
& Def.+CoT        & 94.9  & 97.1  & 8.7   & 33.3 & 91.7  & 16.0  \\
& Def.+Abstain    & 58.5  & 76.7  & 9.6   & 69.9 & 72.1  & 36.2  \\
& Def.+Self-check & 52.4  & 59.7  & 14.4  & 67.6 & 53.8  & 57.7  \\
& Cert.           & 85.0  & 79.2  & 7.7   & 65.1 & 86.9  & 18.6  \\
& Cert.+CoT       & 96.8  & 100.0 & 11.5  & 52.2 & 96.8  & 13.8  \\
\midrule
\multirow{7}{*}{{Haiku}}
& Naive           & 38.0  & 30.4  & 9.3   & 26.6 & 96.2  & 13.1  \\
& Def.            & 97.8  & 99.4  & 10.6  & 57.1 & 82.4  & 27.9  \\
& Def.+CoT        & 100.0 & 99.4  & 15.1  & 75.6 & 92.9  & 22.1  \\
& Def.+Abstain    & 98.4  & 99.0  & 16.3  & 57.1 & 76.6  & 39.7  \\
& Def.+Self-check & 99.0  & 99.4  & 15.7  & 70.5 & 93.3  & 22.4  \\
& Cert.           & 94.6  & 99.7  & 17.6  & 69.6 & 99.4  & 18.3  \\
& Cert.+CoT       & 98.7  & 99.4  & 16.0  & 42.6 & 96.5  & 19.6  \\
\midrule
\multirow{7}{*}{{Gemma}}
& Naive           & 3.8   & 0.0   & 0.0   & 4.5  & 100.0 & 0.0   \\
& Def.            & 49.8  & 42.8  & 0.0   & 40.7 & 100.0 & 0.0   \\
& Def.+CoT        & 99.4  & 99.4  & 6.4   & 61.5 & 98.7  & 6.7   \\
& Def.+Abstain    & 86.6  & 74.1  & 0.0   & 60.3 & 100.0 & 0.0   \\
& Def.+Self-check & 56.2  & 42.2  & 0.0   & 39.4 & 100.0 & 0.0   \\
& Cert.           & 22.4  & 16.9  & 0.0   & 28.8 & 100.0 & 0.0   \\
& Cert.+CoT       & 43.1  & 58.1  & 0.6   & 43.6 & 99.7  & 0.6   \\
\bottomrule
\end{tabular}
\caption{Full coverage-regime results on 1{,}250 complete--partial
pairs (\%). L1--L4 report regime-specific CS. Comp.-CN and Part.-UNK
report the separate member-level correct rates within L3
complete--partial pairs.}
\label{tab:full-partial-regimes}
\end{table}

Table~\ref{tab:full-partial-regimes} extends
Table~\ref{tab:coverage-regimes} to the three additional LLM conditions
omitted from the main-text table, while preserving the same analysis
units and reported columns. Across all seven conditions, L3 remains the
lowest or tied-lowest CS in every model--condition row. Thus, the L3
bottleneck identified for RQ2 is not limited to the four conditions
reported in the main text.

\subsection{Source-Type Family Breakdown of the L3 Asymmetry}

To test whether the L3 gap in
Table~\ref{tab:coverage-regimes} is concentrated in a particular
source-type family, Table~\ref{tab:l3-family} reports member-level
correct rates for each of the four complete and four partial families
under the same four conditions. Each family appears in all five domains,
avoiding a fixed family--domain association. Across all
model--condition cells, complete-family correct rates range from
76.3 to 100.0\%, whereas partial-family correct rates range from
0.0 to 31.6\%. Thus, the RQ2
asymmetry is observed across the full source-type inventory rather than
being concentrated in a single family.

\begin{table}[H]
\centering
\scriptsize
\setlength{\tabcolsep}{5.0pt}\renewcommand{\arraystretch}{0.9}
\begin{tabular}{ll cccc cccc}
\toprule
&& \multicolumn{4}{c}{\textbf{Complete-CN by family}}
& \multicolumn{4}{c}{\textbf{Partial-UNK by family}} \\
\cmidrule(lr){3-6}\cmidrule(lr){7-10}
\textbf{Model} & \textbf{Cond.}
& C1 & C2 & C3 & C4 & P1 & P2 & P3 & P4 \\
\midrule
\multirow{4}{*}{Qwen}
& Naive    & 85.0 & 89.5 & 82.1 & 89.7 & 17.9 & 28.8 & 28.9 & 23.1 \\
& Def.     & 81.2 & 88.2 & 96.2 & 93.6 & 20.5 & 20.0 & 19.7 & 19.2 \\
& Def.+CoT & 90.0 & 85.5 & 94.9 & 96.2 & 9.0  & 18.8 & 22.4 & 14.1 \\
& Cert.    & 88.8 & 81.6 & 79.5 & 97.4 & 17.9 & 22.5 & 18.4 & 15.4 \\
\midrule
\multirow{4}{*}{Haiku}
& Naive    & 95.0 & 97.4 & 94.9 & 97.4 & 11.5 & 16.2 & 13.2 & 11.5 \\
& Def.     & 80.0 & 76.3 & 78.2 & 94.9 & 26.9 & 30.0 & 31.6 & 23.1 \\
& Def.+CoT & 91.2 & 88.2 & 92.3 & 100.0 & 20.5 & 20.0 & 30.3 & 17.9 \\
& Cert.    & 100.0 & 97.4 & 100.0 & 100.0 & 20.5 & 20.0 & 21.1 & 11.5 \\
\midrule
\multirow{4}{*}{Gemma}
& Naive    & 100.0 & 100.0 & 100.0 & 100.0 & 0.0 & 0.0 & 0.0 & 0.0 \\
& Def.     & 100.0 & 100.0 & 100.0 & 100.0 & 0.0 & 0.0 & 0.0 & 0.0 \\
& Def.+CoT & 96.2 & 100.0 & 98.7 & 100.0 & 5.1 & 2.5 & 10.5 & 9.0 \\
& Cert.    & 100.0 & 100.0 & 100.0 & 100.0 & 0.0 & 0.0 & 0.0 & 0.0 \\
\bottomrule
\end{tabular}
\caption{L3 member-level correct rates by source-type family on the
312 complete--partial pairs (\%). C1--C4 are ``system of record,''
``master index,'' ``canonical register,'' and ``definitive index'';
P1--P4 are ``activity feed,'' ``update bulletin,'' ``highlights
digest,'' and ``briefing summary.'' Each family appears in all five
domains; rates are aggregated over domains and over the family used by
the paired member.}
\label{tab:l3-family}
\end{table}

\subsection{Full Prompting Transition Analysis}

Table~\ref{tab:all-transitions} extends
Table~\ref{tab:prompt-shifts} by adding the Abstain and Self-check
transitions omitted from the main-text table and by reporting all
L1--L4 partial groups. The four partial columns together cover the
1{,}250 partial members of the complete--partial branch.
Complete pools the 2{,}500 query-covering complete members, and SM
contains the 1{,}250 Scope-Mismatch members. Each entry is the
after-minus-before accuracy change within the corresponding item group;
positive values indicate accuracy gains.

\begin{table}[H]
\centering
\scriptsize
\setlength{\tabcolsep}{12.5pt}
\renewcommand{\arraystretch}{0.9}
\begin{tabular}{@{}llrrrrrr@{}}
\toprule
\textbf{Model} & \textbf{Transition}
& \textbf{Complete}
& \textbf{L1-Part.}
& \textbf{L2-Part.}
& \textbf{L3-Part.}
& \textbf{L4-Part.}
& \textbf{SM} \\
\midrule
\multirow{4}{*}{\textbf{Qwen}}
& Def.\(\rightarrow\)Def.+CoT
& +8.9 & +34.8 & +13.1 & -3.8 & -7.7 & -11.1 \\
& Def.\(\rightarrow\)Def.+Abstain
& -17.4 & +18.8 & +12.1 & +16.3 & +38.5 & +6.4 \\
& Def.\(\rightarrow\)Def.+Self-check
& -29.1 & +32.3 & +10.2 & +37.8 & +34.9 & +7.5 \\
& Cert.\(\rightarrow\)Cert.+CoT
& +9.9 & +11.8 & +11.2 & -4.8 & -29.8 & -5.7 \\
\midrule
\multirow{4}{*}{\textbf{Haiku}}
& Def.\(\rightarrow\)Def.+CoT
& +11.8 & +2.2 & +0.6 & -5.8 & -4.2 & +3.6 \\
& Def.\(\rightarrow\)Def.+Abstain
& -2.1 & +0.6 & -0.3 & +11.9 & +0.0 & -0.6 \\
& Def.\(\rightarrow\)Def.+Self-check
& +6.6 & +1.3 & +0.0 & -5.4 & -0.6 & -5.4 \\
& Cert.\(\rightarrow\)Cert.+CoT
& +4.4 & +4.2 & +0.0 & +1.3 & -37.2 & -0.6 \\
\midrule
\multirow{4}{*}{\textbf{Gemma}}
& Def.\(\rightarrow\)Def.+CoT
& +0.5 & +49.5 & +56.5 & +6.7 & +6.1 & +9.8 \\
& Def.\(\rightarrow\)Def.+Abstain
& -1.4 & +36.7 & +32.6 & +0.0 & +22.4 & +4.2 \\
& Def.\(\rightarrow\)Def.+Self-check
& +3.4 & +6.4 & -0.6 & +0.0 & -18.3 & -11.4 \\
& Cert.\(\rightarrow\)Cert.+CoT
& +0.3 & +20.8 & +41.5 & +0.6 & +19.9 & +9.9 \\
\bottomrule
\end{tabular}
\caption{Accuracy change after each within-family prompting transition
by item group (percentage points). Complete denotes query-covering
complete members; L1-Part.--L4-Part.\ denote partial members in the
corresponding regimes; SM denotes Scope-Mismatch members. Positive
values indicate accuracy gains.}
\label{tab:all-transitions}
\end{table}

The expanded breakdown shows that the overall results reflect very
different changes across item groups. For Qwen, adding CoT repairs complete,
L1, and L2 cases but causes regressions on L3, L4, and Scope-Mismatch;
Abstain and Self-check improve non-closing cases at a substantial cost
on complete evidence. For Haiku, Definition-aware+CoT improves complete
cases but degrades L3 and L4, while Certificate+CoT sharply degrades L4
with comparatively small changes in the other non-closing groups. Gemma
instead obtains broader gains from CoT, although Self-check regresses on
L4 and Scope-Mismatch. Thus, no transition improves all item groups
consistently across models, supporting the RQ3 conclusion that prompting
redistributes rather than uniformly removes closure errors.

\subsection{Full CROWN-Real Results}

\begin{table}[H]
\centering
\scriptsize
\setlength{\tabcolsep}{4.5pt}
\renewcommand{\arraystretch}{0.9}
\begin{tabular}{@{}ll rrrr rrr@{}}
\toprule
&
& \multicolumn{4}{c}{\textbf{Proceedings}}
& \multicolumn{3}{c}{\textbf{Medical}} \\
\cmidrule(lr){3-6}\cmidrule(lr){7-9}
\textbf{Model} & \textbf{Condition}
& A-CN$\uparrow$
& B-UNK$\uparrow$
& C-UNK$\uparrow$
& \(\Delta_{B-C}\)
& A-CN$\uparrow$
& B-UNK$\uparrow$
& C-UNK$\uparrow$ \\
\midrule

\multirow{5}{*}{Qwen}
& Naive
& 35.0 & 93.6 & 73.7 & 19.9
& 24.7 & 100.0 & 100.0 \\
& Def.
& 0.0 & 100.0 & 100.0 & 0.0
& 8.6 & 100.0 & 100.0 \\
& Def.+CoT
& 88.7 & 99.6 & 66.5 & 33.1
& 94.8 & 100.0 & 97.4 \\
& Cert.
& 93.2 & 100.0 & 91.7 & 8.3
& 71.5 & 98.5 & 100.0 \\
& Cert.+CoT
& 97.4 & 100.0 & 75.2 & 24.8
& 95.5 & 100.0 & 96.3 \\
\midrule

\multirow{5}{*}{Haiku}
& Naive
& 62.8 & 63.2 & 20.3 & 42.9
& 86.5 & 41.2 & 18.7 \\
& Def.
& 99.2 & 32.7 & 3.0 & 29.7
& 99.3 & 24.3 & 19.9 \\
& Def.+CoT
& 100.0 & 98.1 & 27.8 & 70.3
& 100.0 & 100.0 & 85.0 \\
& Cert.
& 95.5 & 100.0 & 12.0 & 88.0
& 91.4 & 100.0 & 95.9 \\
& Cert.+CoT
& 100.0 & 100.0 & 47.4 & 52.6
& 90.6 & 100.0 & 100.0 \\
\midrule

\multirow{5}{*}{Gemma}
& Naive
& 97.0 & 1.1 & 0.8 & 0.4
& 99.3 & 10.5 & 99.3 \\
& Def.
& 99.6 & 0.0 & 0.0 & 0.0
& 100.0 & 41.6 & 98.9 \\
& Def.+CoT
& 100.0 & 93.6 & 3.4 & 90.2
& 98.1 & 100.0 & 100.0 \\
& Cert.
& 100.0 & 98.9 & 1.9 & 97.0
& 97.8 & 100.0 & 99.6 \\
& Cert.+CoT
& 100.0 & 98.9 & 8.3 & 90.6
& 96.3 & 100.0 & 100.0 \\
\bottomrule
\end{tabular}
\caption{Full CROWN-Real variant-level correct rates (\%).
\(\Delta_{B-C}=\text{B-UNK}-\text{C-UNK}\) is reported in percentage
points. Medical C is an explicit-partial control and is not used in
the implicit-partial transfer comparison.}
\label{tab:real-transfer-full}
\end{table}

Table~\ref{tab:real-transfer-full} extends
Table~\ref{tab:real-transfer} with Certificate+CoT and Medical C.
Within Proceedings, B represents narrower-scope completeness and C
represents partial coverage. Across all three models and all five
CROWN-Real conditions, B-UNK is never lower than C-UNK: the difference
is positive in 13 model--condition cells and zero in two, with no
reversal. Thus, the partial-versus-Scope-Mismatch ordering reported for
RQ5 is not limited to the four conditions shown in the main-text table.

Medical C is reported only as an explicit-partial control and is not
used to test implicit-partial transfer. For Medical, both non-closing
variants are generally classified correctly under Definition-aware+CoT,
Certificate, and Certificate+CoT, whereas A-CN remains lower for some
model--condition combinations.

\subsection{Bootstrap Uncertainty Analysis}

Tables~\ref{tab:bootstrap-method}--\ref{tab:bootstrap-real}
report percentile 95\% confidence intervals based on 10{,}000 bootstrap
samples. CROWN-Synth analyses resample matched-pair identifiers from the
relevant subset, whereas the CROWN-Real analysis resamples A/B/C
contrast-set identifiers and retains all three variants of each selected
set. When prompting conditions are compared, both conditions are
evaluated on the same resampled pairs. For the Proceedings B--C
comparison, both variants are evaluated on the same resampled contrast
sets. Branch and regime comparisons resample their corresponding subsets
separately. Confidence intervals including zero do not indicate a clear
directional difference.

\paragraph{Prompting contrasts.}
As shown in Table~\ref{tab:bootstrap-method}, Definition-aware prompting improves Acc and CS for all three models, but
through different directional changes: it reduces both OCR and UCR for
Qwen, while trading large OCR reductions for higher UCR in Haiku and
Gemma. Adding CoT further improves Acc and CS across models, but its OCR
change is inconclusive for Qwen and Haiku and strongly negative for
Gemma. Abstention-aware prompting consistently lowers OCR while
increasing UCR, whereas Self-check remains model-dependent. Certificate
improves Acc and CS for Qwen and Haiku but degrades both for Gemma.
Adding CoT to Certificate improves Acc and CS for Qwen and Gemma, with
no reliable change in either metric for Haiku.

\begin{table}[H]
\centering
\scriptsize
\setlength{\tabcolsep}{10pt}\renewcommand{\arraystretch}{0.9}
\begin{tabular}{ll cccc}
\toprule
\textbf{Model} & \textbf{Contrast} & $\Delta$Acc & $\Delta$OCR & $\Delta$UCR & $\Delta$CS \\
\midrule
\multirow{6}{*}{Qwen} & Def.\,$-$\,Naive & +9.3 {[}8.1, 10.5{]} & -10.3 {[}-12.2, -8.4{]} & -8.3 {[}-10.4, -6.4{]} & +17.2 {[}15.0, 19.5{]} \\
 & Def.+CoT\,$-$\,Def. & +3.9 {[}2.9, 4.9{]} & +1.0 {[}-0.6, 2.6{]} & -8.9 {[}-10.1, -7.6{]} & +8.0 {[}6.1, 9.9{]} \\
 & Def.+Abstain\,$-$\,Def. & -1.7 {[}-2.8, -0.7{]} & -13.9 {[}-15.3, -12.6{]} & +17.4 {[}15.9, 18.9{]} & -3.4 {[}-5.5, -1.4{]} \\
 & Def.+Self-check\,$-$\,Def. & -5.5 {[}-6.6, -4.3{]} & -18.2 {[}-19.7, -16.6{]} & +29.1 {[}27.3, 30.9{]} & -10.6 {[}-12.9, -8.3{]} \\
 & Cert.\,$-$\,Def. & +2.9 {[}1.9, 3.9{]} & -6.6 {[}-8.0, -5.1{]} & +0.7 {[}-0.7, 2.1{]} & +6.0 {[}4.1, 7.9{]} \\
 & Cert.+CoT\,$-$\,Cert. & +2.8 {[}1.9, 3.8{]} & +4.3 {[}2.8, 5.8{]} & -9.9 {[}-11.2, -8.6{]} & +5.8 {[}4.0, 7.6{]} \\
\midrule
\multirow{6}{*}{Haiku} & Def.\,$-$\,Naive & +9.8 {[}8.7, 10.9{]} & -27.5 {[}-29.3, -25.7{]} & +7.8 {[}6.4, 9.2{]} & +19.5 {[}17.3, 21.7{]} \\
 & Def.+CoT\,$-$\,Def. & +6.4 {[}5.6, 7.2{]} & -0.9 {[}-1.9, 0.0{]} & -11.8 {[}-13.2, -10.4{]} & +12.7 {[}11.1, 14.4{]} \\
 & Def.+Abstain\,$-$\,Def. & -0.4 {[}-1.0, 0.2{]} & -1.2 {[}-1.9, -0.5{]} & +2.1 {[}1.1, 3.1{]} & -0.8 {[}-2.0, 0.4{]} \\
 & Def.+Self-check\,$-$\,Def. & +1.6 {[}1.0, 2.3{]} & +3.3 {[}2.5, 4.1{]} & -6.6 {[}-7.8, -5.4{]} & +3.3 {[}2.0, 4.7{]} \\
 & Cert.\,$-$\,Def. & +4.4 {[}3.6, 5.2{]} & 0.0 {[}-1.0, 0.9{]} & -8.7 {[}-10.0, -7.3{]} & +8.7 {[}7.2, 10.3{]} \\
 & Cert.+CoT\,$-$\,Cert. & +0.0 {[}-0.7, 0.7{]} & +4.3 {[}3.2, 5.4{]} & -4.4 {[}-5.3, -3.4{]} & +0.1 {[}-1.2, 1.5{]} \\
\midrule
\multirow{6}{*}{Gemma} & Def.\,$-$\,Naive & +13.5 {[}12.6, 14.5{]} & -31.5 {[}-33.3, -29.6{]} & +4.5 {[}3.7, 5.3{]} & +27.4 {[}25.5, 29.2{]} \\
 & Def.+CoT\,$-$\,Def. & +10.2 {[}9.2, 11.1{]} & -19.8 {[}-21.6, -17.9{]} & -0.5 {[}-1.3, 0.3{]} & +20.4 {[}18.5, 22.2{]} \\
 & Def.+Abstain\,$-$\,Def. & +6.1 {[}5.3, 6.8{]} & -13.6 {[}-15.0, -12.2{]} & +1.4 {[}0.9, 2.0{]} & +11.8 {[}10.4, 13.3{]} \\
 & Def.+Self-check\,$-$\,Def. & -1.9 {[}-2.8, -1.1{]} & +7.3 {[}5.6, 9.0{]} & -3.4 {[}-4.2, -2.7{]} & -3.9 {[}-5.6, -2.2{]} \\
 & Cert.\,$-$\,Def. & -3.8 {[}-4.6, -3.0{]} & +11.0 {[}9.4, 12.7{]} & -3.5 {[}-4.3, -2.7{]} & -7.9 {[}-9.5, -6.3{]} \\
 & Cert.+CoT\,$-$\,Cert. & +7.8 {[}7.0, 8.6{]} & -15.3 {[}-16.9, -13.8{]} & -0.3 {[}-0.9, 0.3{]} & +15.6 {[}14.0, 17.3{]} \\
\bottomrule
\end{tabular}
\caption{Pair-level bootstrap 95\% confidence intervals for the
prompting contrasts (percentage points; 10{,}000 replicates). Negative
\(\Delta\)OCR and \(\Delta\)UCR indicate error reduction. Intervals
excluding zero support a directional difference under pair-level
resampling.}
\label{tab:bootstrap-method}
\end{table}

\paragraph{Partial versus Scope-Mismatch.}
Table~\ref{tab:bootstrap-branch} quantifies uncertainty for the RQ2
branch comparison reported in Table~\ref{tab:pair-decomp} and extended
in Table~\ref{tab:pair-full}. Across every model and condition, the OCR
and Both-CN differences are positive and their intervals exclude zero,
showing consistently greater over-closure on partial evidence than on
Scope-Mismatch evidence. The CS difference is also positive and excludes
zero in most conditions, but is inconclusive for Qwen Self-check and for
Haiku under Definition-aware, Abstain, and Self-check. Thus, although
the pair-discrimination gap weakens under a few interventions, partial
coverage remains the more persistent source of over-closure.

\begin{table}[H]
\centering
\scriptsize
\setlength{\tabcolsep}{10pt}
\renewcommand{\arraystretch}{0.9}
\begin{tabular}{ll ccc}
\toprule
\textbf{Model} & \textbf{Condition}
& \textbf{OCR(part)\,$-$\,OCR(SM)}
& \textbf{CS(SM)\,$-$\,CS(part)}
& \textbf{Both-CN(part)\,$-$\,Both-CN(SM)} \\
\midrule
\multirow{7}{*}{Qwen} & Naive & +42.9 {[}39.5, 46.2{]} & +29.6 {[}25.9, 33.4{]} & +42.0 {[}38.6, 45.4{]} \\
 & Def. & +33.8 {[}30.7, 37.0{]} & +31.0 {[}27.4, 34.6{]} & +34.2 {[}31.1, 37.4{]} \\
 & Def.+CoT & +13.6 {[}10.2, 17.0{]} & +19.0 {[}15.5, 22.6{]} & +13.8 {[}10.4, 17.4{]} \\
 & Def.+Abstain & +18.8 {[}16.3, 21.4{]} & +5.8 {[}1.9, 9.8{]} & +18.6 {[}16.2, 21.1{]} \\
 & Def.+Self-check & +12.6 {[}10.4, 14.7{]} & +1.7 {[}-2.2, 5.6{]} & +12.1 {[}10.0, 14.2{]} \\
 & Cert. & +20.7 {[}17.7, 23.8{]} & +13.6 {[}10.0, 17.3{]} & +20.4 {[}17.4, 23.5{]} \\
 & Cert.+CoT & +17.9 {[}14.6, 21.3{]} & +13.4 {[}9.9, 16.9{]} & +18.6 {[}15.3, 21.9{]} \\
\midrule
\multirow{7}{*}{Haiku} & Naive & +54.2 {[}51.1, 57.5{]} & +43.0 {[}39.4, 46.5{]} & +54.6 {[}51.5, 57.9{]} \\
 & Def. & +10.8 {[}8.2, 13.4{]} & +1.7 {[}-2.1, 5.4{]} & +10.7 {[}8.1, 13.4{]} \\
 & Def.+CoT & +16.2 {[}13.7, 18.6{]} & +14.5 {[}11.4, 17.6{]} & +16.2 {[}13.7, 18.6{]} \\
 & Def.+Abstain & +7.1 {[}4.6, 9.7{]} & -3.0 {[}-6.8, 0.7{]} & +7.3 {[}4.8, 9.8{]} \\
 & Def.+Self-check & +6.6 {[}3.8, 9.6{]} & -1.6 {[}-5.2, 2.0{]} & +6.8 {[}3.9, 9.8{]} \\
 & Cert. & +17.4 {[}14.9, 20.0{]} & +10.8 {[}7.5, 14.2{]} & +17.4 {[}14.8, 19.9{]} \\
 & Cert.+CoT & +24.7 {[}21.9, 27.6{]} & +23.4 {[}20.2, 26.6{]} & +24.9 {[}22.1, 27.8{]} \\
\midrule
\multirow{7}{*}{Gemma} & Naive & +43.8 {[}40.9, 46.6{]} & +42.2 {[}39.4, 45.0{]} & +44.0 {[}41.2, 46.9{]} \\
 & Def. & +35.7 {[}32.0, 39.4{]} & +34.3 {[}30.6, 38.0{]} & +35.5 {[}31.9, 39.2{]} \\
 & Def.+CoT & +15.8 {[}12.5, 19.1{]} & +8.3 {[}4.8, 11.8{]} & +15.0 {[}11.8, 18.3{]} \\
 & Def.+Abstain & +16.9 {[}13.4, 20.5{]} & +14.1 {[}10.3, 17.8{]} & +17.2 {[}13.7, 20.8{]} \\
 & Def.+Self-check & +27.4 {[}23.5, 31.1{]} & +24.2 {[}20.4, 28.1{]} & +27.8 {[}23.9, 31.5{]} \\
 & Cert. & +54.4 {[}51.1, 57.7{]} & +51.2 {[}47.8, 54.6{]} & +54.6 {[}51.3, 57.9{]} \\
 & Cert.+CoT & +43.6 {[}40.2, 47.0{]} & +43.8 {[}40.3, 47.2{]} & +43.6 {[}40.2, 47.0{]} \\
\bottomrule
\end{tabular}
\caption{Bootstrap 95\% confidence intervals for partial versus
Scope-Mismatch branch differences (percentage points). Positive OCR
and Both-CN differences indicate more over-closure on partial evidence;
a positive CS difference indicates lower pair discrimination on the
partial branch. }
\label{tab:bootstrap-branch}
\end{table}

\paragraph{Coverage-regime contrasts.}
Table~\ref{tab:bootstrap-regime} quantifies uncertainty for the
regime-level and L3 member-level contrasts underlying the RQ2 results in
Table~\ref{tab:coverage-regimes}.
L3 has lower CS than L1 in every model--condition cell and lower CS than
L2 in every cell except the exact L2--L3 tie for Gemma under Naive
prompting; all nonzero L1 and L2 contrasts exclude zero. The L3 CS point
estimate is also lower than L4 in every condition, although the Qwen
Naive interval includes zero. The L3 complete-CN minus partial-UNK gap
is positive and excludes zero in every condition except Qwen
Self-check. Thus, the bootstrap analysis supports the RQ2 conclusion
that L3 is the most persistent regime-level bottleneck and that its
failure is usually one-sided, concentrated on implicit-partial members.

\begin{table}[H]
\centering
\scriptsize
\setlength{\tabcolsep}{10pt}\renewcommand{\arraystretch}{0.9}
\begin{tabular}{ll cccc}
\toprule
\textbf{Model} & \textbf{Condition}
& \textbf{CS(L1)\,$-$\,CS(L3)}
& \textbf{CS(L2)\,$-$\,CS(L3)}
& \textbf{CS(L4)\,$-$\,CS(L3)}
& \textbf{L3 cCN\,$-$\,pUNK} \\
\midrule
\multirow{7}{*}{Qwen} & Naive & +32.9 {[}26.2, 39.6{]} & +24.2 {[}17.8, 30.6{]} & +4.8 {[}-0.6, 10.3{]} & +61.9 {[}54.2, 69.6{]} \\
 & Def. & +49.2 {[}42.8, 55.6{]} & +60.4 {[}54.3, 66.4{]} & +30.1 {[}23.7, 36.5{]} & +69.9 {[}62.8, 76.6{]} \\
 & Def.+CoT & +86.2 {[}82.1, 90.1{]} & +88.5 {[}84.6, 92.0{]} & +24.7 {[}18.6, 30.8{]} & +75.6 {[}69.2, 81.7{]} \\
 & Def.+Abstain & +48.9 {[}42.5, 55.2{]} & +67.1 {[}61.3, 72.8{]} & +60.3 {[}53.8, 66.3{]} & +35.9 {[}26.3, 45.2{]} \\
 & Def.+Self-check & +38.0 {[}31.3, 44.7{]} & +45.3 {[}38.6, 52.0{]} & +53.2 {[}46.5, 59.6{]} & -3.8 {[}-13.8, 6.1{]} \\
 & Cert. & +77.3 {[}72.2, 82.1{]} & +71.5 {[}66.1, 77.0{]} & +57.4 {[}51.3, 63.1{]} & +68.3 {[}61.2, 75.3{]} \\
 & Cert.+CoT & +85.3 {[}81.1, 89.1{]} & +88.5 {[}84.9, 91.7{]} & +40.7 {[}34.3, 47.1{]} & +83.0 {[}78.2, 87.5{]} \\
\midrule
\multirow{7}{*}{Haiku} & Naive & +28.7 {[}22.3, 34.8{]} & +21.1 {[}15.0, 27.1{]} & +17.3 {[}11.5, 23.1{]} & +83.0 {[}77.9, 87.8{]} \\
 & Def. & +87.2 {[}83.3, 91.0{]} & +88.8 {[}85.3, 92.3{]} & +46.5 {[}40.1, 52.9{]} & +54.5 {[}45.8, 62.8{]} \\
 & Def.+CoT & +84.9 {[}80.8, 88.8{]} & +84.3 {[}80.1, 88.1{]} & +60.6 {[}54.5, 66.3{]} & +70.8 {[}64.1, 77.2{]} \\
 & Def.+Abstain & +82.1 {[}77.6, 86.5{]} & +82.7 {[}78.5, 86.9{]} & +40.7 {[}34.0, 47.4{]} & +36.9 {[}27.6, 45.8{]} \\
 & Def.+Self-check & +83.3 {[}78.9, 87.2{]} & +83.7 {[}79.5, 87.5{]} & +54.8 {[}48.1, 60.9{]} & +70.8 {[}64.4, 77.2{]} \\
 & Cert. & +76.9 {[}71.8, 81.7{]} & +82.1 {[}77.9, 86.2{]} & +51.9 {[}45.2, 58.3{]} & +81.1 {[}76.6, 85.6{]} \\
 & Cert.+CoT & +82.7 {[}78.5, 86.9{]} & +83.3 {[}79.2, 87.5{]} & +26.6 {[}19.9, 33.3{]} & +76.9 {[}71.5, 82.4{]} \\
\midrule
\multirow{7}{*}{Gemma} & Naive & +3.8 {[}1.9, 6.1{]} & +0.0 {[}0.0, 0.0{]} & +4.5 {[}2.2, 6.7{]} & +100.0 {[}100.0, 100.0{]} \\
 & Def. & +49.8 {[}44.4, 55.3{]} & +42.8 {[}37.4, 48.2{]} & +40.7 {[}35.3, 46.2{]} & +100.0 {[}100.0, 100.0{]} \\
 & Def.+CoT & +93.0 {[}90.1, 95.5{]} & +93.0 {[}90.1, 95.8{]} & +55.1 {[}49.0, 60.9{]} & +92.0 {[}88.8, 94.9{]} \\
 & Def.+Abstain & +86.6 {[}82.7, 90.4{]} & +74.1 {[}69.3, 78.9{]} & +60.3 {[}54.8, 65.4{]} & +100.0 {[}100.0, 100.0{]} \\
 & Def.+Self-check & +56.2 {[}50.8, 61.7{]} & +42.2 {[}36.7, 47.6{]} & +39.4 {[}34.0, 44.9{]} & +100.0 {[}100.0, 100.0{]} \\
 & Cert. & +22.4 {[}17.9, 27.2{]} & +16.9 {[}12.8, 21.1{]} & +28.8 {[}23.7, 34.0{]} & +100.0 {[}100.0, 100.0{]} \\
 & Cert.+CoT & +42.5 {[}37.1, 47.9{]} & +57.5 {[}51.8, 62.9{]} & +42.9 {[}37.5, 48.4{]} & +99.0 {[}97.8, 100.0{]} \\
\bottomrule
\end{tabular}
\caption{Bootstrap 95\% confidence intervals for coverage-regime
contrasts on the 1{,}250 complete--partial pairs (percentage points).
The first three columns report
\(\mathrm{CS}(Lk)-\mathrm{CS}(L3)\); positive values indicate lower CS
for L3 than for the compared regime. The final column reports the L3
complete-CN minus partial-UNK correct-rate gap.}
\label{tab:bootstrap-regime}
\end{table}

\paragraph{CROWN-Real transfer contrast.}
For Proceedings, we resample the 266 contrast-set identifiers with
replacement and retain the A, B, and C variants of each selected set.
Table~\ref{tab:bootstrap-real} reports bootstrap intervals for
\(\Delta_{B-C}=\mathrm{B\mbox{-}UNK}-\mathrm{C\mbox{-}UNK}\).
A positive difference indicates that partial C is more error-prone than
narrower-complete B.

\begin{table}[H]
\centering
\scriptsize
\setlength{\tabcolsep}{12pt}
\renewcommand{\arraystretch}{0.9}
\begin{tabular}{l ccc}
\toprule
\textbf{Condition}
& \textbf{Qwen}
& \textbf{Haiku}
& \textbf{Gemma} \\
\midrule
Naive
& +19.9 {[}13.9, 25.9{]}
& +42.9 {[}36.1, 49.2{]}
& +0.4 {[}-1.1, 1.9{]} \\
Def.
& +0.0 {[}0.0, 0.0{]}
& +29.7 {[}23.7, 35.7{]}
& +0.0 {[}0.0, 0.0{]} \\
Def.+CoT
& +33.1 {[}27.4, 38.7{]}
& +70.3 {[}64.7, 75.9{]}
& +90.2 {[}86.5, 93.6{]} \\
Cert.
& +8.3 {[}5.3, 11.7{]}
& +88.0 {[}83.8, 91.7{]}
& +97.0 {[}94.7, 98.9{]} \\
Cert.+CoT
& +24.8 {[}19.9, 30.1{]}
& +52.6 {[}46.6, 58.6{]}
& +90.6 {[}86.8, 94.0{]} \\
\bottomrule
\end{tabular}
\caption{Contrast-set bootstrap 95\% confidence intervals for
\(\Delta_{B-C}=\mathrm{B\mbox{-}UNK}-\mathrm{C\mbox{-}UNK}\)
on the 266 Proceedings A/B/C sets
(percentage points; 10{,}000 replicates).
Positive values indicate that partial C is more error-prone than
narrower-complete B. The A/B/C variants of each selected set are retained
together. Confidence intervals including zero do not indicate a clear
directional difference.}
\label{tab:bootstrap-real}
\end{table}

Across the 15 model--condition cells, all point estimates of
\(\Delta_{B-C}\) are nonnegative, and the confidence intervals exclude
zero in 12 cells. Of the remaining three, two are exact ties under
Definition-aware prompting: B-UNK and C-UNK are both 100.0\% for Qwen
and both 0.0\% for Gemma. The third is a near-tie for Gemma under Naive
prompting. Thus, partial C is more error-prone with a confidence interval
excluding zero in 12 cells, while the remaining three show no clear
directional difference. No observed point estimate is negative.

\end{document}